\documentclass[conference]{IEEEtran}
\usepackage{times}
\usepackage{tabularx,ragged2e}
\usepackage[numbers]{natbib}
\usepackage{multicol}
\usepackage[bookmarks=true]{hyperref}
\usepackage{multirow}
\usepackage{array}
\usepackage{textgreek}
\newcolumntype{C}{>{\arraybackslash}X}

\usepackage{amsmath,amssymb,stmaryrd,mathtools}
\usepackage{xcolor}
\usepackage{xspace}
\usepackage{bbm}
\usepackage{graphicx}
\usepackage{subfig} 
\definecolor{dark_red}{RGB}{122, 0, 0}
\definecolor{coral}{RGB}{255, 119, 94}
\definecolor{pink_orange}{RGB}{255, 72, 126}
\definecolor{vibrant_pink}{RGB}{255, 0, 104}
\definecolor{pink_pink}{RGB}{255, 37, 153}
\definecolor{wine}{RGB}{204, 0, 102}

\definecolor{light_orange}{RGB}{255, 198, 107}
\definecolor{orange(sae/ece)}{rgb}{1.0, 0.49, 0.0}
\definecolor{dark_orange}{RGB}{216,92,0}

\definecolor{org-purp-0}{RGB}{165, 76, 0}
\definecolor{org-purp-1}{RGB}{250, 130, 28}
\definecolor{org-purp-2}{RGB}{226, 89, 68}
\definecolor{org-purp-3}{RGB}{206, 92, 124}
\definecolor{org-purp-4}{RGB}{116, 80, 146}
\definecolor{org-purp-5}{RGB}{110, 78, 157}

\definecolor{teal(sae/ece)}{rgb}{0, 0.47, 0.52}
\definecolor{aqua}{RGB}{52,172,139}
\definecolor{dark_aqua}{RGB}{35,115,93}
\definecolor{dark_green}{RGB}{0, 92, 34}

\definecolor{grape}{RGB}{112,48,160}
\definecolor{purple}{rgb}{0.74, 0.65, 1.0}
\definecolor{dark_purple}{rgb}{0.58, 0.0, 0.82}
\definecolor{periwinkle}{RGB}{191, 140, 230}

\definecolor{light_gray}{rgb}{0.9, 0.9, 0.9}
\definecolor{medium_gray}{rgb}{0.6, 0.6, 0.6} 
\definecolor{dark_gray}{rgb}{0.2, 0.2, 0.2} 

\definecolor{sky_blue}{RGB}{37, 166, 213}
\definecolor{light_blue}{rgb}{0.33, 0.80, 1}
\definecolor{dark_blue}{rgb}{0.098, 0.239, 0.52}
\definecolor{ocean}{RGB}{13, 121, 202}
\definecolor{light_ocean}{RGB}{18, 178, 235}
\definecolor{dark_ocean}{RGB}{10, 89, 148}
\definecolor{vibrant_blue}{RGB}{14, 120, 255}

\definecolor{dark_brown}{rgb}{0.3255, 0.004, 0.001}

\newcommand{\para}[1]{\medskip\noindent\textbf{#1. }}

\newcounter{qnum}
\newcounter{tnum}
\newcommand{\ours}{\textcolor{org-purp-1}{\textbf{FOREWARN}}\xspace}
\newcommand{\oursoracle}{\textcolor{org-purp-0}{\textbf{FOREWARN-Oracle}}\xspace}

\newcommand{\vlmimgoracle}{\textcolor{dark_blue}{\textbf{VLM-Img-Oracle}}\xspace}
\newcommand{\vlmimg}{\textcolor{ocean}{\textbf{VLM-Img}}\xspace}
\newcommand{\vlmact}{\textcolor{sky_blue}{\textbf{VLM-Act}}\xspace}
\newcommand{\classdynlatent}{\textcolor{org-purp-4}{\textbf{Classifier-Dyn-Latent}}\xspace}
\newcommand{\vlmdynlatentbin}{\textcolor{org-purp-3}{\textbf{VLM-DynLat-Binary}}\xspace}
\newcommand{\vlmdynlatentcat}{\textcolor{org-purp-3}{\textbf{VLM-DynLat-Category}}\xspace}
\newcommand{\basepolicy}{\textcolor{black}{\textbf{Base Policy}}\xspace}

\newcommand{\obs}{o}
\newcommand{\img}{I}
\newcommand{\proprio}{q}
\newcommand{\imgSpace}{\mathcal{I}}
\newcommand{\proprioSpace}{\mathcal{Q}}
\newcommand{\obstraj}{\mathbf{o}}
\newcommand{\obsSpace}{\mathcal{O}}

\newcommand{\enc}{\mathcal{E}_\phi}
\newcommand{\dec}{\mathcal{D}_\phi}
\newcommand{\dyn}{f_\phi}

\newcommand{\lang}{\ell}
\newcommand{\langSpace}{\mathcal{L}}

\newcommand{\vlm}{\text{VLM}}
\newcommand{\behavior}{\ell_b}

\newcommand{\latent}{z}
\newcommand{\latenttraj}{\mathbf{z}}
\newcommand{\latentSpace}{\mathcal{Z}}

\newcommand{\action}{a}
\newcommand{\acttraj}{\mathbf{\action}}

\newcommand{\thor}{T}

\newcommand{\rewvlm}{R^{\vlm}}

\usepackage{amsmath}

\begin{document}

\title{From Foresight to Forethought: VLM-In-the-Loop Policy Steering via Latent Alignment}

\author{Yilin Wu$^{1}$, Ran Tian$^{2}$, Gokul Swamy$^{1}$, Andrea Bajcsy$^{1}$\\
$^{1}$Carnegie Mellon University $^{2}$UC Berkeley
\\
\{yilinwu, gswamy, abajcsy\}@andrew.cmu.edu, 
rantian@berkeley.edu
}

\makeatletter
\let\@oldmaketitle\@maketitle%
\renewcommand{\@maketitle}{\@oldmaketitle%
\setcounter{figure}{0} %
\centering
\includegraphics[width=1.0\textwidth]{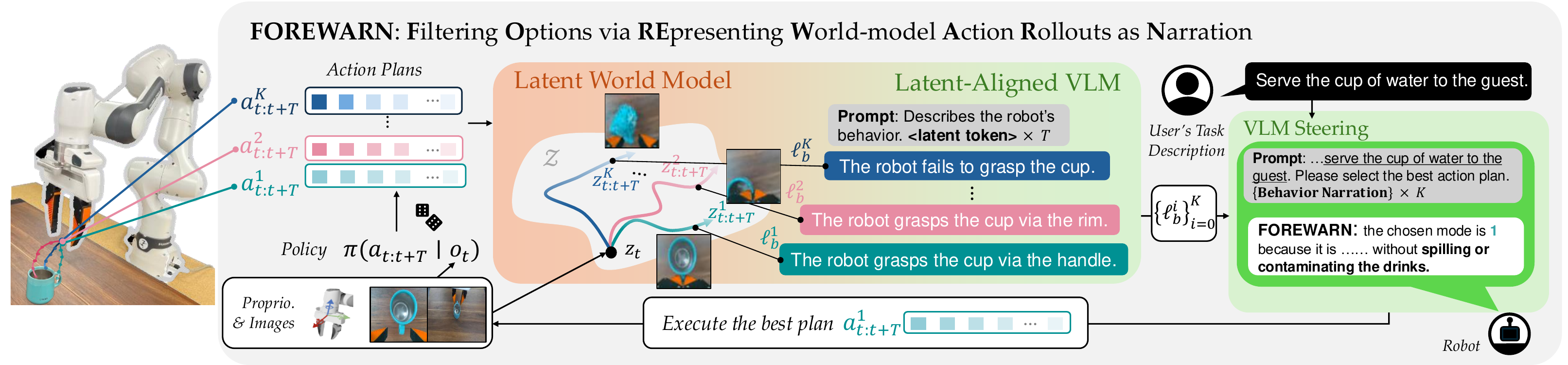}
\captionof{figure}{We present \textbf{FOREWARN}, an VLM-in-the-loop policy steering algorithm for multi-modal generative robot policies. 
Our key idea is to decouple the VLM's burden of predicting action outcomes from evaluation. 
By predicting action outcomes with a pre-trained latent dynamics model and aligning a VLM to reason about these latent states in text, FOREWARN can select action plans at runtime that are most appropriate for new task contexts and user needs. 
}
\label{fig:front-fig}
\vspace{-0.3in}
\bigskip}
\makeatother
\maketitle

\begin{abstract}
While generative robot policies have demonstrated significant potential in learning complex, multimodal behaviors from demonstrations, they still exhibit diverse failures at deployment-time. 
Policy steering offers an elegant solution to reducing the chance of failure by using an external verifier to select from low-level actions proposed by an imperfect generative policy. 
Here, one might hope to use a Vision Language Model (VLM) as a verifier, leveraging its open-world reasoning capabilities. 
However, off-the-shelf VLMs struggle to understand the consequences of low-level robot actions as they are represented fundamentally differently than the text and images the VLM was trained on. 
In response, we propose FOREWARN,
a novel framework to unlock the potential of VLMs 
as open-vocabulary verifiers for runtime policy steering. 
Our key idea is to decouple the VLM's burden of predicting action outcomes (\textit{foresight}) from evaluation (\textit{forethought}). 
For foresight, we leverage a latent world model to imagine future latent states given diverse low-level action plans. 
For forethought, we align the VLM with these predicted latent states to 
reason about the consequences of actions in its native representation---natural language---and effectively filter proposed plans. 
We validate our framework across diverse robotic manipulation tasks, demonstrating its ability to bridge representational gaps and provide robust, generalizable policy steering. Videos can be found on the project website: ~\href{https://yilin-wu98.github.io/forewarn/}{https://yilin-wu98.github.io/forewarn/}.
\end{abstract}

\IEEEpeerreviewmaketitle

\section{Introduction}
Generative imitation-based policies are an increasingly powerful way to learn low-level robot behaviors from multimodal\footnote{Here, multimodality only refers to the training data's action distribution.} expert demonstrations~\citep{chi2024diffusionpolicy, fu2024mobile,zhao2023learningfinegrainedbimanualmanipulation}. 
Despite their impressive ability to learn diverse behaviors directly from high-dimensional observations, these policies still degrade in nuanced and unexpected ways at runtime.  
For example, consider the robot in the left of Figure~\ref{fig:front-fig} that must pick up a mug from the table. 
At training time, the generative policy learns a distribution over useful interaction modes such as grasping the cup by different parts (e.g., handle, lip and interior, etc.) shown in wrist camera photo in Figure~\ref{fig:front-fig}.

However, at runtime, the policy exhibits a range of degradations, from complete task failures (such as the robot knocking down the cup during grasping, shown in 
the center of Figure~\ref{fig:front-fig}), to inappropriate behaviors that are misaligned with the deployment context or preferences of an end-user (such as the robot placing its gripper inside of a cup of water when serving a guest shown in the middle of Figure~\ref{fig:front-fig}). 
While a common mitigation strategy involves re-training the policy via more demonstrations~\citep{shafiullah2024supervised} or interventions~\citep{ross2010efficient, liumulti},  runtime failures are not always an indication that the base policy is fundamentally incapable of producing the desired behavior. 
In fact, the base policy may already contain the ``right'' behavior mode within its distribution (e.g., grasping the cup by the handle), but due to putting too much probability mass on an undesired mode, the robot does not reliably choose the correct action plan at runtime. 

Runtime policy steering \citep{nakamoto2024steering,wang2024inference} offers an elegant solution to this mode-selection problem. 
By using an external \textit{verifier}
to select candidate plans proposed by an imperfect generative policy, the robot's behavior can be ``steered'' at runtime without any need for re-training.
Despite the initial successes demonstrated by the policy steering paradigm, the question still remains of how to fully unlock autonomous policy steering in the open world when the robot's environment, task context, and base policy's performance are constantly changing.

Policy steering can be approached via the stochastic model-predictive control framework of modern control theory, which decomposes the optimal action selection (i.e. \textit{generation}) problem of runtime policy steering into \textit{(a)} \textit{predicting} the outcomes of a given action plan and \textit{(b)} \textit{verifying} how well they align with user intent. However, this approach is only feasible when a physics-based, low-dimensional dynamics model is available for outcome prediction and a well-defined reward function can be specified for verification. In open-world environments, both of these requirements are challenging to fulfill due to the complexity of dynamics modeling and the difficulty of hand-crafting rewards to evaluate nuanced task requirements \citep{hadfield2017inverse}.

Our core idea is to leverage world models, which are well-suited for predicting action outcomes in open world settings, and VLMs, which have great potential as verifiers due to their commonsense reasoning abilities, to develop a divide-and-conquer approach to open-world policy steering. However, doing so naively is challenging as world models and VLMs operate on fundamentally different representations of reality.

To address this concern, we propose  \textbf{FOREWARN}: \textbf{F}iltering \textbf{O}ptions via \textbf{RE}presenting \textbf{W}orld-model \textbf{A}ction \textbf{R}ollouts via \textbf{N}arration.
To predict challenging action outcomes (e.g., interaction dynamics of a manipulator and a deformable bag), we use state-of-the-art world models
\citep{liumulti,wu2023daydreamer} to predict lower-dimensional latent state representations from high-dimensional observation-action data (shown in orange in the center of Figure~\ref{fig:front-fig}). 
To critique behavior outcomes under nuanced task specifications (e.g., ``Serve the cup of water to the guest''), we leverage vision-language models (VLMs)~\cite{dubey2024llama,openai2024gpt4technicalreport} as our open-world verifiers (shown in green in the center of Figure~\ref{fig:front-fig}).
Importantly, we demonstrate that \textit{aligning} the VLM to reason directly about the predicted latent states from the world model enables it to understand fine-grained outcomes that it cannot directly predict zero-shot nor understand from image reconstructions. 
Ultimately, this alignment step enables our ``VLM-in-the-loop'' policy steering approach to interpret action plans as behavior narrations and select high-quality plans by reasoning over those narrations even under novel task descriptions
(shown on the right of Figure~\ref{fig:front-fig}).

We evaluate \textbf{FOREWARN} on a suite of robotic manipulation tasks, demonstrating how it can robustly filter proposed action plans to match user intent and task goals even when faced with variations not seen during training. In summary, our main contributions are:
\begin{itemize}
    \item Formalizing runtime policy steering a stochastic model-predictive control problem, revealing the \textit{generation-verification gap} \citep{Godel, cook2023complexity, swamy2025all} and where state-of-the-art models have maximal potential to shine. 
    \item A latent space alignment strategy that enables a VLM to more reliably verify action plan outcomes predicted by a low-level, action-conditioned world model. 
    \item A novel, fully-autonomous policy steering framework that improves a base generative imitation-based policy by over $30\%$, even in novel task descriptions. 
    \item Extensive experiments in hardware showing that our latent-aligned VLM approach outperforms (by $\sim40\%$) altnerative VLM approaches that do \textit{not} decouple the prediction of outcomes from verification. 
\end{itemize}

\section{Related Work}

\para{Generative Imitation-Based Policies}
With the rise of large-scale open-source datasets of expert demonstrations~\citep{droid, rt12022arxiv,  rt22023arxiv,contributors2024agibotworldrepo, fang2023rh20t, open_x_embodiment_rt_x_2023}, imitation learning (IL) has become a popular way to learn low-level robot control policies from data. 
In particular, recent advances in generative modeling have unlocked policy architectures that can model diverse, multi-modal behaviors directly from high-dimensional observations~\citep{chi2024diffusionpolicy,leebehavior,reuss2024multimodal}. 
At the same time, generative IL policies still exhibit nuanced, hard-to-anticipate performance degradations during deployment time.  
These degradations range from complete task failures (e.g., inability to grasp a cup, knocking it down, or dropping it~\citep{nakamoto2024steering}) potentially due to distribution shifts or visual distractors~\citep{hancock2024run,vincent2024generalizable}, to inappropriate behaviors that are misaligned with the deployment context or an end-user's preferences (e.g., placing the gripper inside of a cup filled with water during grasping)~\citep{agiaunpacking}. 
In this work, our goal is to leverage the diverse low-level behavior distribution that the base policy has learned, but prevent these nuanced performance degradations at runtime via our novel policy steering method.

\para{Failure Detection, Monitoring \& Prediction}
The handling of generative policy failures can be grouped into three broad categories: \textit{posthoc} detection, \textit{runtime} monitoring, and failure \textit{prediction}. 
\textit{Posthoc} approaches identify and explain failures present in offline robot datasets, and have recently leveraged Vision Language Models (VLMs) to accelerate this process via video captioning, highlighting critical data frames, and providing human-interpretable summaries of failures ~\citep{duanaha, guan2024task,liu2023reflect, wangcan}. 
In contrast, \textit{runtime} monitoring aims to detect failures as they happen during robot deployment.
To quickly identify nuanced failures, recent methods propose a ``fast and slow'' approach: a fast online detector flags unusual situations (e.g., binary anomaly classifier), while a slower VLM-based reasoner provides a deeper understanding of the event and if it is a relevant failure~\citep{agiaunpacking,SinhaElhafsiEtAl2024}. 
Although these strategies can effectively identify failures, they fundamentally require the robot to start failing for the runtime monitor to activate. 
The final category, failure \textit{prediction} methods, anticipate failures before they occur and unlock the potential for preemptive correction of the base policy. 
Here, existing approaches~\citep{kambara2025futuresuccesspredictionopenvocabulary,liu2023modelbased,liumulti} often rely on out-of-distribution (OOD) detection in a latent space or dense human labels to train a binary classifier that distinguishes failures from successes. 
In this work, we contribute to the \textit{predictive} category of methods. Our method anticipates future outcomes of the policy's actions via a latent world model, and reasons about the outcomes via a VLM that is aligned with the predicted latent states. 

\para{Policy Steering}
A traditional method to improve a base IL policy is to fine-tune it with additional intervention data~\citep{liumulti} or recovery behaviors~\citep{shafiullah2024supervised}. 
However, recently, runtime policy steering has become an attractive alternative to improving a generalist IL policy \cite{nakamoto2024steering, wang2024inference} without needing any additional and expensive demonstration data.  
Runtime policy steering assumes that the base policy is capable of generating the correct behaviors, but fails to select them reliably. 
Here, an external verifier can be used to re-rank (i.e., ``steer'') the generations towards ones with good outcomes. 
Previous methods have explored humans-in-the-loop~\citep{wang2024inference} or a $Q$-function learned from very large offline datasets labeled with sparse rewards. ~\citep{nakamoto2024steering} as the verifier. In our framework, by using a VLM-in-the-loop as our verifier, we can perform policy steering autonomously after finetuning VLM on a small dataset and provide human-like interpretable guidance.

\para{Learning to Search} From an algorithmic perspective, our approach fits within the paradigm of \textit{learning to search} (L2S) \citep{ratliff2009learning}. In L2S, one learns the components required to, at test time, plan a sequence of actions, rather than merely imitating what was in the training data. L2S provides two key advantages over direct imitation algorithms like behavioral cloning. First, the agent gains the ability to reason about the consequences of its actions and potentially recover from mistakes, avoiding \textit{compounding errors} \citep{ross2011reduction, swamy2021moments}. We see this manifest in our system's ability to avoid or correct from failures the base policy would have produced otherwise. Second, \textit{verifying} whether a plan is good is often easier to learn than \textit{generating} a good plan in the first place \citep{swamy2025all, ng2000algorithms}. We see this manifest in the fact that our system only requires a limited amount of data to fine-tune our verifier VLM, rather than the larger amount of data an end-to-end approach would have required \citep{shalev2016sample}.

In contrast to more classical L2S approaches that require extensive \textit{global} search in the real world \citep{ratliff2009learning, ziebart2008maximum}, we instead perform \textit{local} search \citep{bagnell2003policy} against a learned verifier \citep{swamy2023inverse, espinosa2025efficient} inside a learned world model \citep{ren2024hybrid}. This allows us to avoid the computational expense of global search and the potential safety violations incurred by real-world interaction. As argued by \citep{swamy2023inverse, ren2024hybrid}, doing so still matches many of the guarantees of classical L2S methods if one fits the world model on a mixture of on-policy and off-policy trajectories (i.e., in a \textit{hybrid} fashion \citep{song2022hybrid, vemula2023virtues}), as we do consistently across all of our experiments.

\section{Problem Formulation}
\label{sec:problem-formulation}

We formalize the general problem of policy steering as a stochastic model-predictive control problem over the set of action plans proposed by a base action generation model. 

\begin{figure*}[t]
    \centering
    \includegraphics[width=\linewidth]{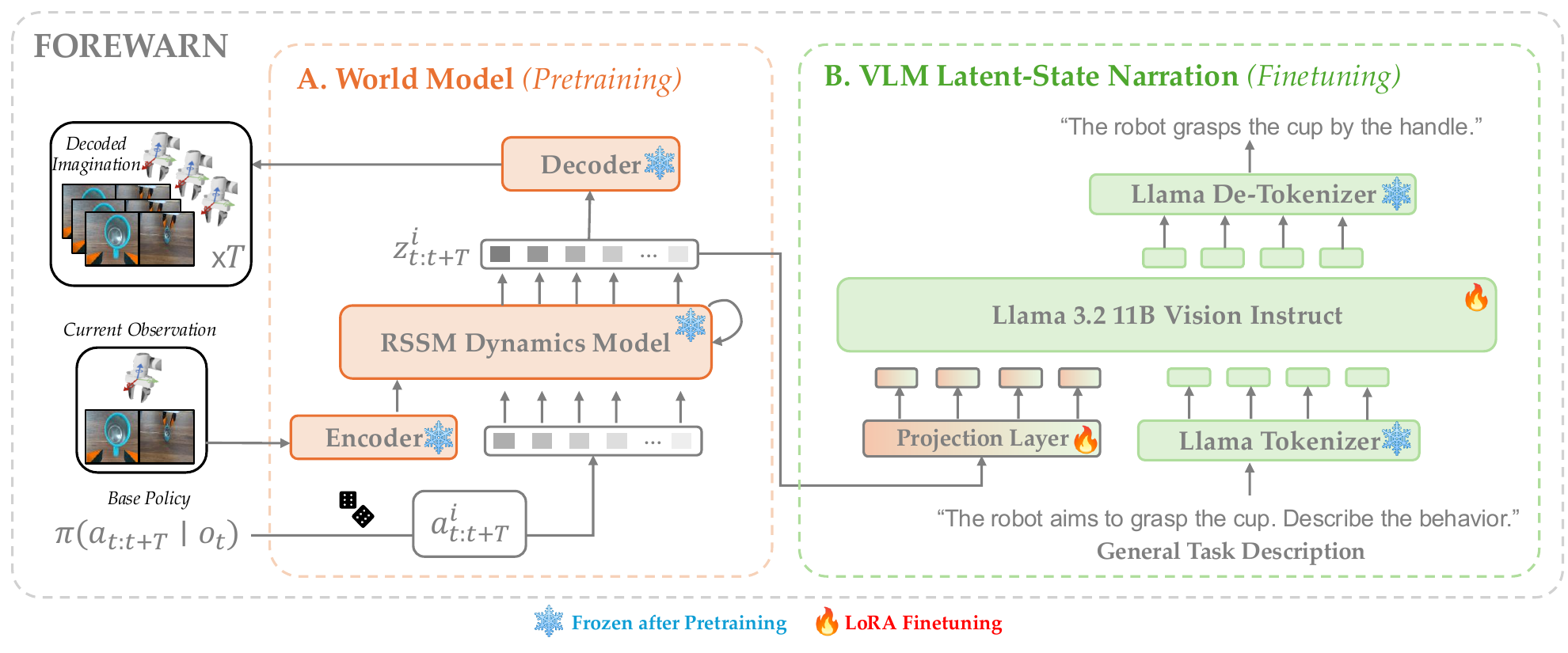}
    \caption{\textbf{Training FOREWARN.} In part A (Sec.~\ref{sec:foresight}), a Recurrent State Space Model (RSSM) is pretrained to learn good latent embeddings of the dynamics conditioned on the observations and actions. In part B (Sec.~\ref{sec:forethought}), the sequence of learned latent embeddings is projected through a linear layer to the text embedding space, similar to the original vision token processing in the Llama-3.2 Model. The projection layer and Llama model are finetuned together using LoRA \cite{hu2022lora}, but the world model remains frozen. }
    \label{fig:overview_fig}
\end{figure*}

\para{Setup \& Notation} 
Let the robot's pre-trained multimodal imitative action generation model \citep{chi2024diffusionpolicy,leebehavior} be denoted by  $\pi(\acttraj_t\mid \obs_t)$. We will often refer to this model as the robot's ``base policy'' throughout this paper with which it performs a task for an end-user. 
The robot’s observations $\obs \in \obsSpace := \imgSpace \times \proprioSpace$ combine RGB image data $\img \in \imgSpace$ and proprioceptive states $\proprio \in \proprioSpace$ (e.g., end-effector pose, gripper state), and $\acttraj_t := \action_{t:t+\thor}$ denotes a robot's $\thor$ step action plan, with each action in the sequence specifying end-effector positions and rotations. Similarly, $\obstraj_t := \obs_{t:t+\thor}$ is an observation sequence. In the real world, after observing some $\obs_t$, generating some action plan $\acttraj_t \sim \pi(\cdot \mid \obs_t)$ and executing it open-loop, the robot observes a sequence of observations $\obstraj_t \sim \mathcal{P}(\cdot| \obs_t, \acttraj_t)$.

\para{Problem} Given the robot's current observation $\obs_t$ at timestep $t$ and $K$ \textit{i.i.d.} samples from the base policy, $\{\acttraj^i_t\}^K_{i=1} \sim \pi(\acttraj_t\mid \obs_t)$, the \textit{policy steering} problem seeks to return the action plan $\acttraj^{\star}_t$ which optimizes the following objective:  
\begin{align}
    \acttraj^{\star}_t = \arg\max_{\acttraj_t \in \{\acttraj^i_t\}^K_{i=1}} \mathbb{E}_{\obstraj_t\sim \mathcal{P}(\cdot \mid \obs_t, \acttraj_t) } \Big[R(\obstraj_t;\lang )\Big].
    \label{eq:policy-steering-oc-problem}
\end{align} 
Fundamentally, solving the \textit{(behavior) generation} problem specified in Eq.\ref{eq:policy-steering-oc-problem} requires two abilities: \textit{prediction} and \textit{verification}. 
The \textit{prediction} problem can be clearly seen in the expectation of Eq.\ref{eq:policy-steering-oc-problem}, wherein we have to forward simulate the outcomes of any action plan $\acttraj^i_t$ and ``imagine'' the potential future observations, $\obstraj_t \sim \mathcal{P}(\obstraj_t \mid \obs_t, \acttraj^i_t)$. 
The \textit{verification} problem lies inside of the expectation and with the reward function $R(\obstraj_t; \lang)$. 
Here, $\lang \in \langSpace$ is the task description, represented via a language description (e.g., ``Serve the cup of water to the guest.''), and $R$ verifies how well or how poorly the future observations align with the behavior specified by $\ell$.

While Eq.\ref{eq:policy-steering-oc-problem} characterizes the underlying policy steering problem, it is extremely challenging to solve end-to-end because of the coupled prediction and verification steps. 
This is where we seek to leverage vision-language models in-the-loop to obtain a practical solution with the potential to generalize to new environments and hard-to-model steering criteria.
Initially, it may be tempting use the VLM directly as a black-box solver of Eq.\ref{eq:policy-steering-oc-problem} (i.e. to solve the overarching behavior generation problem) by simply passing it the $K$ action plan options, the current observation $\obs_t$, and the task description $\lang$, and having it directly return $\acttraj^{\star}_t$. 
However, low-level action data is beyond the training distribution of current VLMs, which primarily focus on high-level semantic understanding and not embodied control~\citep{hu2023toward}. 
Alternatively, we could fine-tune the VLM to directly select action plans via labeled observation-action-samples datasets (labeled with optimal action trajectories). 
However, this strategy is sample-inefficient, requiring extensive embodied rollouts and human annotations to generate labels. 
Instead, we propose tackling the problem in Eq.\ref{eq:policy-steering-oc-problem} in a way that leverages the unique strengths of a VLM. 
We hypothesize that the right place to put VLMs into the loop is specifically in the \textit{verification} step above, as it leverages the model’s strong reasoning abilities \textit{given predicted outcomes}.
Then, to complement the VLM, we propose that \textit{prediction} should be handled by an embodied world model that can reason about hard-to-model outcomes directly from low-level actions.
It is well known that verification is significantly easier than generation for many problems \citep{cook2023complexity, swamy2025all}, boding well for the sample-efficiency of our modular approach.

\section{Our Approach: FOREWARN}
\label{sec:method}
Our key idea is to adopt a divide-and-conquer strategy, explicitly decoupling the VLM's burden of predicting action outcomes from evaluation during policy steering. 
Specifically, we take advantage of recent advances in latent dynamics models \citep{liumulti,wu2023daydreamer} which can learn lower-dimensional latent state representations from high-dimensional observation-action data collected on a robot. 
By passing possible action generations to the latent dynamics model, we can efficiently predict diverse future outcomes that would hard to model otherwise. 
Importantly, even though the world model learns a latent state that is an approximate sufficient statistic of the dynamics, one still needs to teach the VLM to evaluate outcomes \textit{in the latent representation of the world model}, rather than from the raw image observations, as we will address in detail below.

At the highest level, our mathematical formulation of model-predictive ``VLM-in-the-loop'' policy steering is: 
\begin{equation}
    \begin{aligned}
    \acttraj^{\star}_t &= \arg\max_{\acttraj_t \in \{\acttraj^i_t\}^K_{i=1}} \mathbb{E}_{\latenttraj_t \sim \dyn(\latent_t, \acttraj_t)} \Big[R^\vlm_\psi\big(\latenttraj_t; \lang \big) \Big], \\ 
    & \text{s.t.} \quad\quad \latent_t = \enc(\obs_t),\\ 
\end{aligned}
\label{eq:vlm_mpc_latent}
\end{equation}
where $\latent_{t+1} \sim \dyn(\latent_t, \action_t)$ is a probabilistic latent dynamics model, $\latenttraj_{t}:=\latent_{t:t+\thor}$, $\enc: \obsSpace \rightarrow \latentSpace$ is an observation encoder that maps the robot’s current observations $\obs_t$ into a latent space $\latentSpace$, and $R^
\vlm_\psi(\latenttraj_t; \lang)$ represents an implicit reward function embedded in our %
VLM (parameterized by $\psi$) which evaluates the predicted outcomes given the task description. Both $f_\phi$ and $\enc$ are part of our world model and parameterized by $\phi$. In short, to realize our idealized policy steering objective \eqref{eq:policy-steering-oc-problem}, we approximate the expectation over future outcomes with world model rollouts and leverage a fine-tuned VLM as a latent-outcome-conditioned verifier, leading to \eqref{eq:vlm_mpc_latent}.
We call our overall policy steering approach \textbf{FOREWARN}: \textbf{F}iltering \textbf{O}ptions via \textbf{RE}presenting \textbf{W}orld-model \textbf{A}ction \textbf{R}ollouts as \textbf{N}arration, visualize it in \autoref{fig:front-fig}, and now discuss the details.

\subsection{\textit{Foresight}: 
Predicting Outcomes via Latent World Models} 
\label{sec:foresight}

In this work, we adopt the Dreamerv3 world model ~\citep{hafner2023mastering} as our mechanism for predicting future outcomes of action plans. 
This world model (visualized on the left in Fig.~\ref{fig:overview_fig}) is pre-trained via an offline dataset that contains trajectories of the robot's observations and actions: $\mathbb{D}_{\text{WM}} = \{\{ (\obs_\tau^j, \action_\tau^j)\}_{\tau=0}^{H-1}\}_{j=1}^M$, where $M$ is the total number of trajectories with each trajectory representing the robot's outcome $\obs_{0:H}^j$ induced by an action plan $\action_{0:H}^j$, and $H$ denotes the length of these trajectories. 
The training data consists of both successful and failed rollouts from the base policy $\pi(\acttraj \mid \obs)$ and additional demonstration data. This allows the world model to accurately predict the outcomes of both good and bad actions plans, a sufficient condition for successful behavior generation \citep{ren2024hybrid}.

The world model training loss incentivizes the latent state to be informative for high-quality image decoding as well as highly predictive of the next latent state given an action (see \cite{hafner2023mastering} and Appendix~\ref{sec:appendix_world_model} for more details). 
After training, the world model is frozen, and we utilize the trained $\enc$ and $\dyn$ throughout the rest of our policy steering algorithm.

\subsection{\textit{Forethought}: Latent-Text Alignment for Outcome Reasoning and Policy Steering}
\label{sec:forethought}

Once the world model encoder $\enc$ learns an effective low-dimensional latent state representation $\latent_t$ and the latent dynamics model $\dyn$ learns to predict future latent states conditioned on the robot’s actions, we can shift our focus to evaluating the imagined outcomes of any candidate low-level action plan. 
We hypothesize that the VLM’s open-world reasoning capabilities could allow it to be an effective verifier here, enabling better decision-making downstream.

However, before we can unlock the potential of the VLM as a verifier, we need to first enable the VLM to reason directly about the predicted latent states (rather than in terms of raw image observations). We propose approaching this problem as a \textit{latent-text alignment} problem: by mapping latent states generated by the world model to a textual representation, we can enable the VLM to more easily evaluate plans due to its strong natural language understanding abilities. 
We then further frame this alignment problem as a \textit{visual question-answering (VQA)} task, where we prompt the VLM to narrate the real-world behavior of the robot conditioned on a sequence of latent states produced by the world model (visualized on the right of \autoref{fig:overview_fig}). We now discuss this process in detail.

\para{VLM Backbone}
\label{para:vlm model}
We use the Llama-3.2-11B-Vision-Instruct model~\citep{dubey2024llama} as our VLM backbone. The original Llama model %
utilizes a Vision Transformer (ViT) to tokenize visual inputs into latent tokens, which are then processed by the LLM backbone to generate text outputs.
In our setting, we adapt the Llama model by replacing its observation (image) tokenization module with our world model’s encoder $\enc$ and latent dynamics model $\dyn$. To align the latent dynamics embedding space with the text embedding space, we introduce a linear layer as an adapter. 
Specifically, $\enc$ encodes the current robot observation, and the forward dynamics model $\dyn$ predicts future latent states based on a candidate action plan (left side of \autoref{fig:overview_fig}). These latent states are passed through the linear layer to produce a sequence of latent tokens, which serve as input to the LLM backbone (right side of \autoref{fig:overview_fig}).

\para{VLM Finetuning} Our objective is to enable the VLM to narrate the real-world behavior of the robot, denoted as $\behavior^i$, based on a sequence of generated latent states, $\latent^i_{t:t+\thor}$ from the world model. These behavior narrations highlight fine-grained motion details (e.g., ``the robot grasps the cup by the handle") rather than the high-level semantics of the motion (e.g., ``the robot grasps the cup"). This design encourages the model to capture nuanced behaviors and identify potential failures in the robot’s predicted outcomes. We construct a VQA dataset to finetune the VLM to achieve this objective (described in Sec.~\ref{sec:experiments} and App.~\ref{sec:appendix_vlm}).

\para{VLM-In-the-Loop Policy Steering} 
Our fine-tuned VLM can now describe the outcomes of candidate action plan in natural language narration. However, our ultimate goal is to query the VLM to identify the best action sequence. To accomplish this, we propose querying the same VLM again (i.e. using it as a verifier), leveraging its open-world reasoning capabilities and natural language understanding to select the best action plan.
Mathematically, policy steering can be expressed as:
\begin{equation}
\begin{aligned}
    \acttraj^{\star}_t &= \arg\max_{\acttraj_t \in \{\acttraj^i_t\}^K_{i=1}} \mathbb{E}_{\latenttraj_t \sim \dyn(\latent_t, \acttraj_t)} \Big[\rewvlm_\psi \big(\mathcal{T}^{\vlm}_\psi(\latenttraj_t; \lang); \lang \big) \Big], \\  
    & \text{s.t.} \quad\quad \latent_t = \enc(\obs_t).
   \label{eq:vlm_reward}
    \end{aligned}
\end{equation}
where $\mathcal{T}^{\vlm}_\psi : \latentSpace^\thor \rightarrow \mathcal{L}_{b}$ denotes the inference process that translates a candidate action plan into the associated robot behavior narration. Instead of instructing the VLM to explicitly assign the reward for each predicted action plan outcome, we leverage its implicit knowledge about the reward and directly query the VLM for the best action plan among the $K$ candidates, conditioned on the task description $l$ (see the example in right side of \autoref{fig:front-fig}), which can be seen as a form of multiple-choice question answering (MCQA).

In summary, by integrating the predictions from the world model with the VLM’s open-vocabulary behavior narration generation and commonsense reasoning, \textit{FOREWARN} guides the robot towards action plans that are aligned with the deployment context and task goals and enables the robot to proactively prevent failures.

\section{Experiments}
\label{sec:experiments}
In this section, we instantiate several real-world manipulation tasks to study our method. 
We first investigate how effectively we can translate low-level actions into high-level behavior descriptions (Sec.~\ref{sec:behavior}).%
Then we evaluate the closed-loop policy steering performance as well as our method's robustness to novel task descriptions, $\lang$ (Sec.~\ref{sec:steering}).

\para{Real Robot Setup} We use a Franka Emika robotic manipulator equipped with a 3D printed gripper from~\citep{chi2024universal} in our experiments. 
The base generative policy controls the low-level robot actions (which are the robot's end-effector pose and gripper opening) controlled at 15Hz.
The robot uses RGB images from a wrist-mounted camera and a third-person camera mounted in front of the robot. See App.~\ref{sec:appendix_robot_setup} for more details.

\para{Manipulation Tasks} We consider three real-world robot manipulation tasks that exhibit underlying multi-modal behaviors, hard-to-model outcomes, and nuanced failures. 
In the \textbf{Cup} task, the robot must grasp a cup placed in front of it and in the \textbf{Bag} task, the robot must pick up a bag of chips from the table. 
Fundamentally, both tasks can be accomplished in diverse ways: for example, for the \textbf{Cup} task, the robot can pick up the cup by the handle or by placing its fingers inside the cup by grasping the lip; for the \textbf{Bag} task, the robot can grab the bag by any edge, or by squeezing the center. 
Furthermore, the \textbf{Bag} task has more challenging dynamics and potential outcomes because the bag is deformable. 
We use this task to study how our framework performs when faced with harder-to-predict interaction outcomes and nuanced failures (e.g., crushing the chips inside the bag). 

\begin{figure*}[t!]
    \centering
    \includegraphics[width=\linewidth]{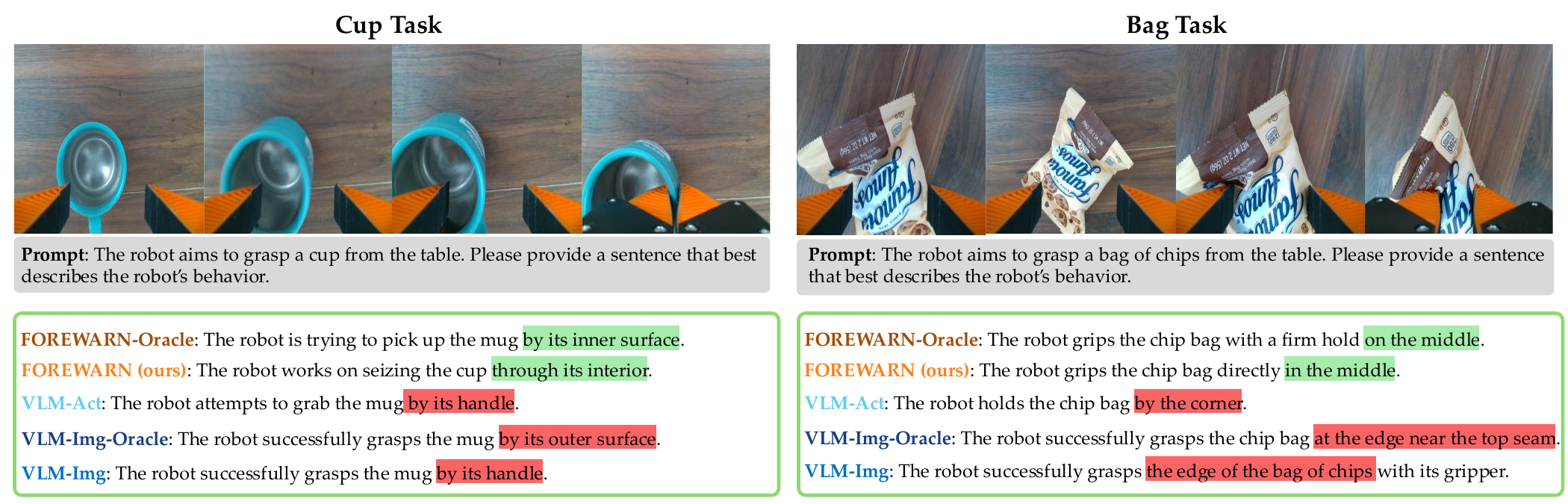}
    \caption{\textbf{Examples of Behavior Narrations Predicted by Each Approach.} The top row displays the ground-truth robot observations and the prompt used for querying VLMs.
Only \ours and \oursoracle consistently produce accurate outcome narrations, effectively capturing nuanced motion details. In contrast, the baselines frequently hallucinate or fail to capture critical contact details between the gripper and objects. For instance, in the \textbf{Bag} task, \vlmact, \vlmimg, and \vlmimgoracle all hallucinate that the robot is grasping the edge of the bag, whereas it is actually grasping the middle.}
    \label{fig:behavior_qualitative}
\end{figure*}

We also introduce a third task that features longer horizon and more complex interactions \textbf{\textit{Fork-to-Bowl Transfer}}.
In this task, the robot must pick up a fork from the table and place it inside a bowl. 
This task is considerably more challenging than the other two tasks for three reasons: 1) it requires longer-horizon planning to effectively navigate distinct phases; 2) requires reasoning about interactions between objects 
(the fork and the bowl); and 3) it introduces different multi-modal motion choices, including selecting the optimal picking location (tines or handle) and determining the appropriate placement strategy (inside versus outside the bowl, low versus high release).
In the top row of Figure~\ref{fig:behavior_qualitative}, we visualize the camera observations of the robot interacting with these items for Bag and Cup tasks.

\para{Base Policy} For our multi-modal imitative action generation model, $\pi(\acttraj_t \mid \obs_t)$, we use a Diffusion Policy~\citep{chi2024diffusionpolicy} trained on 100 teleoperated demonstrations per task. 
The policy takes inputs from wrist and third-person cameras, along with proprioceptive states, to predict a distribution over $\thor$-step action plan, where $\thor = 64$. Please see App.~\ref{sec:appendix_base_policy} for more details. %

\para{World Model Training}We collected 250 real-world trajectories per task, including both successful and failed rollouts from the base policy, along with additional 100 demonstrations used in base policy, for training and evaluating the world model (300 for training and 50 for testing).
The world model is trained to predict $\thor=64$ future latent states 
given the current $\obs_t$ and an action plan $\acttraj_t$. %

\para{VLM Fine-tuning} We construct our VQA dataset for fine-tuning from the same offline dataset, $\mathbb{D}_{\text{WM}}$, used to train the world model. 
For each $\thor$-step trajectory snippet  $ \{(\obs_\tau^j, \action_\tau^j\}_{\tau=t}^{t+\thor}$ from the dataset, the encoder $\enc$ processes the initial observation $\obs_t^j$ at timestep $t$, and the forward dynamics model $\dyn$ predicts latent states $\latent^j_{t:t+\thor}$ based on the action plan $\action^j_{t:t+\thor}$. We note that in our specific implementation, although $\dyn$ is a stochastic dynamics function, we use only the most likely prediction as the outcome associated with the action plan. To avoid ``semantically repetitive'' latent states between adjacent low-level actions, we downsample the T-step latent states to T/4.  
These downsampled latent states, along with the task description $\ell$, are provided as input to the VLM, and we manually annotate the corresponding behavior narrations, $\behavior \in \langSpace_{b}$, for the associated observations $\obs_{t:t+\thor}^j$.
We fine-tune the model using the Low-Rank Adaptation (LoRA) technique~\citep{hu2022lora}, keeping both the encoder $\enc$ and the latent dynamics model $\dyn$ frozen during the fine-tuning process.

\para{VLM-In-the-Loop Policy Steering} When deploying \ours for run-time policy steering, we begin by sampling 100 action plans from the base policy and aggregating them into $K=6$ modes using the non-maximum suppression (NMS) scheme from \cite{seff2023motionlm}. These 6 aggregated action plans are then passed to \ours for interpretation and evaluation. 
For each candidate action plan, we use only the most likely future latent state predictions from $\dyn$ as input to the VLM for reasoning.

\subsection{From Action Rollouts to Behavior Narration}\label{sec:behavior}

As discussed in Sec.~\ref{sec:forethought}, if we want to use the VLM as an open world verifier $\rewvlm_\psi(\cdot; \lang)$, we need to enable the model to understand the underlying textual representation of low-level action outcomes.
In this section, we study if our our latent-aligned VLM can accurately describe the outcomes of low-level actions. 
We also compare our approach with several baselines to investigate the advantages of using an explicit world model for predicting action outcomes and decoding a robot’s action plans into behavior narrations.

\para{Baselines} We compare our approach, \ours, against four baselines (more implementation details provided in App.~\ref{sec:appendix_additional_baselines}). 
(1) \oursoracle, is an upper-bound on our method's performance assuming that we had access to \textit{ground-truth} future observations (instead of relying on the latent dynamics $\dyn$ to predict future outcomes). This method uses the encoder $\enc$ on ground-truth future observations to get privileged (posterior) future latent states $\latent_{t:t+\thor}$ as input for the VLM. %
(2) \vlmact, which directly fine-tunes the original Llama-3.2-11B-Vision-Instruct model to generate behavior narrations end-to-end from the current observation $\obs_t$ and an action plan $\action_{t:t+\thor}$ (represented as text), without explicitly predicting outcomes. %
(3) \vlmimg, which utilizes the \textit{decoded} world model's predictions (i.e., the predicted future visual observations) given a robot’s planned actions. We use GPT-4o~\citep{openai2024gpt4technicalreport} to process the predicted visual observations and generate behavior narrations in a zero-shot manner.
(4) \vlmimgoracle, which is similar to \vlmimg but is an upper bound on this method by using ground-truth visual observations instead of predicted ones.

\para{Metrics} %
We adopt the metrics from~\citep{duanaha} to evaluate the alignment between predicted behavior narrations and ground-truth narrations:
(1) \textbf{LLM Score}: A similarity score (ranging from $0$ to $1$) determined by the GPT-4o model.
(2) \textbf{GT Accuracy}: A binary score ($0$ or $1$) indicating whether the predictions match the ground-truth narrations, as determined by a human labeler (in this case, the authors). For further details on the motivation behind using these metrics for evaluation, please refer to App.~\ref{sec:appendix_ablations}.

  \begin{table}[ht]
  \vspace{-0.3cm}
   \tiny
        \centering
        \setlength{\tabcolsep}{4pt}
        \renewcommand{\arraystretch}{1.3}
        \begin{tabular}{c|ccc|ccc}
        \hline
           \multirow{2}{*}{} & \multicolumn{3}{c|}{\textbf{GT Accuracy} $\uparrow$ } & \multicolumn{3}{c}{\textbf{LLM Score} $\uparrow$}\\
           \cline{2-7}
        & \textbf{Cup} & \textbf{Bag} & Average & \textbf{Cup} & \textbf{Bag} & Average\\
            \hline
            \oursoracle & 0.92$\pm$0.02 & 0.77$\pm$0.03 & 0.85$\pm$0.03 &  0.86$\pm$0.01 & 0.76$\pm$0.03 & 0.81$\pm$0.02 \\
            \hline
            \ours (Ours) & \textbf{0.87}$\pm$\textbf{0.02} & \textbf{0.75}$ \pm $\textbf{0.03} & \textbf{0.82}$\pm$\textbf{0.03} & \textbf{0.82}$\pm$\textbf{0.02} & \textbf{0.72}$\pm$\textbf{ 0.02}&\textbf{0.76}$\pm$\textbf{0.02} \\
            \vlmact & 0.37$\pm$0.03 & 0.36$\pm$0.06 & 0.37$\pm$0.05 & 0.52$\pm$0.05 & 0.50$\pm$0.07 & 0.51$\pm$0.06\\
            \vlmimgoracle & 0.61$\pm$0.04 & 0.43$\pm$0.03 & 0.52$\pm$0.04 & 0.65$\pm$0.02 & 0.62$\pm$0.02 & 0.64$\pm$0.02\\
            \vlmimg &  0.36$\pm$0.05 & 0.33$\pm$0.04& 0.35$\pm$0.05 & 0.56$\pm$0.03 & 0.60$\pm $0.04 & 0.58$\pm$0.04\\ 
            \hline 
        \end{tabular}
        \caption{\textbf{Alignment Between Predicted Behavior Narrations and Ground-Truth Narrations.} \ours outperforms all baselines across both tasks and achieves performance comparable to \oursoracle, which has access to ground-truth action outcomes and represents the upper bound for our approach.  We use 50 rollouts to evaluate the performance. For \ours, \oursoracle and \vlmact, the mean and standard deviation are reported by running 3 seeds for the finetuning experiments while \vlmimg and \vlmimgoracle, report 3 queries of GPT-4o. }
        \label{tab:vlm-ablation_all} 
        \vspace{-0.3cm}
    \end{table}

\para{Results: On the Value of Explicit Action Outcome Prediction} Table~\ref{tab:vlm-ablation_all} presents the \textbf{GT Accuracy} and \textbf{LLM Score} for our approach and each baseline. The results are averaged across 30 test rollouts for each task.
The results show that \vlmact performs poorly, achieving less than $50\%$ \textbf{GT Accuracy} across all tasks. This underperformance is due to its inability to interpret low-level actions without the grounding provided by a world model’s future outcome predictions. 
In contrast, \ours, which leverages an explicit world model, outperforms \vlmact by over $50\%$ on every task, despite both being fine-tuned on the same dataset. These results demonstrate that decoupling the VLM’s burden of predicting action outcomes enables the model to produce more accurate outcome narrations than directly training the VLM to both predict outcomes and generate narrations end-to-end.

\para{Results: On the Value of VLM Fine-Tuning}
Interestingly, despite being among the most advanced VLMs, both \vlmimg and \vlmimgoracle struggle to accurately interpret robot behaviors directly from visual observations, even with access to ground-truth observations. As shown in Table~\ref{tab:vlm-ablation_all}, these methods fall behind \ours by at least $30\%$ in \textbf{GT Accuracy} and $16\%$ in \textbf{LLM Score}. These results show that existing state-of-the-art VLMs struggle to decode fine-grained motion details from video observations, underscoring the importance of fine-tuning for improved performance in such tasks.

\para{Results: Qualitative Examples}
In \autoref{fig:behavior_qualitative}, we visualize behavior narrations generated by our approach and the baselines. \ours consistently produces more accurate outcome narrations, effectively capturing nuanced motion details. In contrast, the baselines often hallucinate or fail to capture critical contact details between the gripper and objects.

\begin{figure}[t!]
    \centering
    \vspace{-0.3cm}\includegraphics[width=1\linewidth]{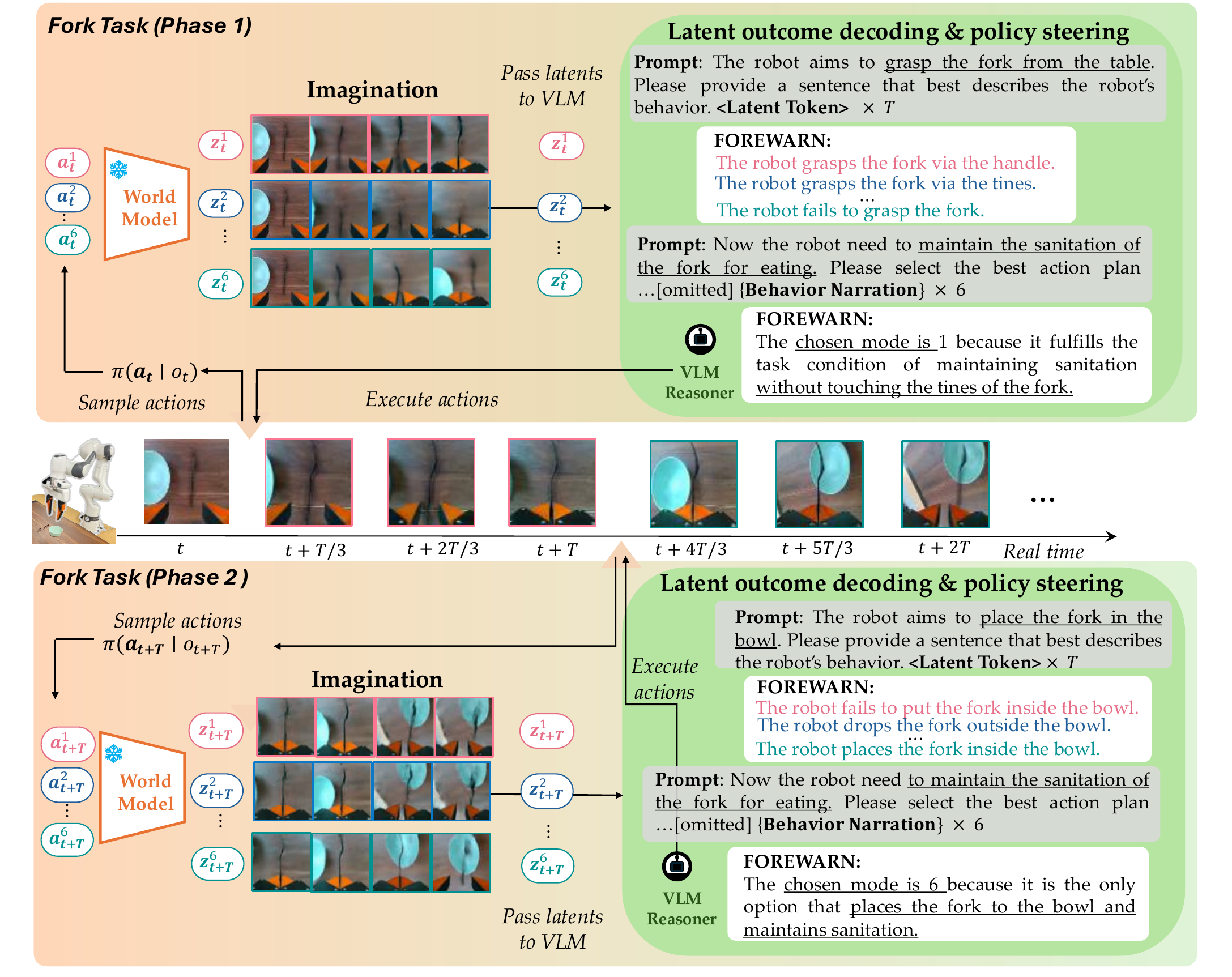}
    \caption{\textbf{Policy Steering: Fork Task.} We visualize the steering process for the \textbf{Fork} task including two phases (Pick and Place).  For each phase, we visualize the imagined $\thor$-step rollouts decoded from the world model for the 3 out of 6 action plans sampled from the base policy on the left. On the right, we show the behavior narrations generated from our finetuned VLM $\mathcal{T}^{\vlm}_\psi$ and the VLM's reasoning $\rewvlm_\psi(\cdot; \lang)$ about the outcomes based on the task description $\lang$ and behavior narrations to select the best action plan to execute. The time axis shows the real-world execution of the selected behavior from the perspective of the wrist-camera.}
    \label{fig:policy-steering}
\end{figure}

\subsection{Policy Steering for Open-World Alignment}
\label{sec:steering}

The results from the previous section demonstrate our approach's effectiveness in decoding predicted latent states into nuanced behavior narrations by explicitly using a world model for outcome prediction.
In this section, we compare our approach against several baselines to evaluate its system-level policy steering performance and robustness under \textbf{novel} task descriptions and specifications. Our experiments focus on evaluating the impact of leveraging the VLM as both an interpreter and evaluator of predicted action outcomes.

\para{Baselines} 
We first keep the VLM’s role as the verifier unchanged and ablate the effect of using an explicit world model to predict action outcomes. Specifically, we compare \ours to \vlmact (described in \autoref{sec:behavior}), which directly fine-tunes the original Llama model to predict action outcome narrations without utilizing a world model. Similar to our approach, \vlmact then queries the VLM to select the best motion plan based on the task description $l$.
Next, we keep the world model unchanged and ablate the effect of alignment: mapping latent states generated by the world model to their underlying textual representation that the VLM can easily reason about.

We compare our approach against two more baselines: (3) \vlmdynlatentcat, which also leverages a world model for outcome predictions and a VLM for action plan selection. However, instead of decoding the predicted outcomes into text representations and leveraging the VLM's open-world knowledge for policy steering, it directly fine-tunes the VLM to predict a set of indices of the successful candidate action plans and randomly selects one index from the set to execute the corresponding action plan.
(4) \classdynlatent, which is similar to \vlmdynlatentcat, but instead of relying on a VLM, it directly takes the predicted latent embeddings $\hat{\latent}_{t:t+\thor}$ as input and trains a transformer-based binary classifier (commonly used in prior failure-prediction work \citep{liumulti}) to predict success or failure for each candidate action plan and randomly selects one predicted to succeed.

\para{Metrics} We evaluate each policy steering method using the \textbf{Success Rate}. For each method, we conduct 20 trials with randomly initialized task configurations and report the average success rate across these trials. A trial is considered successful if the robot successfully completes the task while aligning with the end-user’s preferences.

     \begin{table}[ht]
        \centering
        \tiny
        \setlength{\tabcolsep}{3.5pt}
        \renewcommand{\arraystretch}{1.5}
        \begin{tabular}{c|ccc| ccc}
        \hline
            \multirow{3}{*}{Method} & \multicolumn{6}{c}{Success Rate $\uparrow$}\\
            \cline{2-7}
            &\multicolumn{3}{c|}{Training Task Description} & \multicolumn{3}{c}{Novel Task Description} \\
            \cline{2-7}
            & Cup & Bag & Fork & Cup & Bag & Fork \\
            \hline 
            \basepolicy & 0.30$\pm$0.14 & 0.20$\pm$0.13 & 0.10$\pm$0.09 & 0.50$\pm$0.16 & 0.40$\pm$0.15 & 0.30$\pm$0.14 \\ \hline 
            \ours (Ours) & \textbf{0.80$\pm$0.13} & \textbf{0.70$\pm$0.14} & \textbf{0.70$\pm$0.14} & \textbf{0.80$\pm$0.13} & \textbf{0.70$\pm$0.14} & \textbf{0.60$\pm$0.15}\\
            \vlmdynlatentcat & \textbf{0.80$\pm$0.13} & 0.40$\pm$0.15 & 0.50$\pm$0.16  & 0.30$\pm$0.14 & 0.40$\pm$0.15  & 0.30$\pm$0.14 \\
            \classdynlatent & \textbf{0.80$\pm$0.13} & \textbf{0.70$\pm$0.14} & \textbf{0.70$\pm$0.14} & 0.00$\pm$0.00 & 0.10$\pm$0.09 & 0.20$\pm$0.13\\
            \vlmact & 0.40$\pm$0.15 & 0.20$\pm$0.13 & 0.20$\pm$0.13 & 0.30$\pm$0.14 & 0.50$\pm$0.16 & 0.20$\pm$0.13\\
            \hline
        \end{tabular}
        \caption{\textbf{Policy Steering.} The success rate is reported by averaging over 20 different rollouts. \ours outperforms all the baselines in both training and novel task contexts. }
        \label{tab:policy_steering_all}
    \end{table}

\para{Results: Policy Steering Performance} 
We first evaluate our approach and the baselines under task descriptions $\ell$ that fall within the training distribution.
Table~\ref{tab:policy_steering_all} shows the success rates of the base robot policy (without policy steering), our approach (\ours), and the baselines for each task.
Our results demonstrate that \ours can effectively steer the policy towards safe and aligned behavior modes by leverging the VLM as an interpreter and evaluator of predicted latent action outcomes.

Interestingly, \classdynlatent achieves comparable performance to our approach in all tasks under this setting while \vlmdynlatentcat fails to match \ours in more complicated \textbf{Bag} and \textbf{Fork} tasks. \vlmact has similar performance as base policy because the behavior narrations, directly generated from low-level action sequences, fail to capture accurate motion details and thus provide no useful signal for VLM to steer the policy.    Next, we assess their generalization abilities under novel task descriptions to evaluate the robustness of our approach.

\para{Results: Robustness to Novel Task Descriptions} 
We evaluate the success rate of each approach when the task description is altered to introduce novel scenarios.
Specifically, in the \textbf{Cup} task, we modify the original task description from \textit{“Please grasp a cup from the table and serve the cup of water to the guest”} to a novel description: \textit{“Please grasp a cup from the table, but note that the handle is covered with oil.”} This change makes behaviors where the robot grasps the cup by the handle unsafe, while grasping the cup by the rim becomes the desired behavior.
In the \textbf{Bag} task, we modify the original task description from \textit{“Please pick up a bag of chips from the table and minimize the contact region to avoid crushing contents inside”} to a novel description: \textit{“Please pick up a bag of chips from the table and maximize stability without dropping the bag.”} This change makes behaviors where the robot picks up the bag by the corner less preferred, as the bag may slip from the gripper, while squeezing the middle of the bag to secure it becomes the desired behavior.
In the \textbf{Fork} task, we modify the original prompt from \textit{"Please pick up the fork and place it in the bowl while maintaining the sanitation for eating."} to a novel description: \textit{“Please pick up the fork and place it in the bowl while maximizing the contact region during grasping.”} This change makes the behaviors where the robot picks up the handle and drop the fork high less preferred as the fork handle is more narrow than tines and dropping high can make fork fall out of the bowl while grasping by tines is more secure. 

Interestingly, while both \vlmdynlatentcat and \classdynlatent improve task success rates when task descriptions fall within the training distribution, their performance deteriorates significantly with novel task descriptions. This indicates that the VLM struggles to reason directly about predicted action outcomes from the world model’s latent states and essentially degrades to a traditional end-to-end model. To fully leverage the VLM’s open-world reasoning capabilities for generalized policy steering, it is essential to enable the model to interpret predicted action outcomes through textual representations.

\para{Results: Qualitative Examples}
In \autoref{fig:policy-steering}, we present examples of runtime policy steering using our approach for the \textbf{Fork} task and additional examples for \textbf{Cup} and \textbf{Bag} tasks are included in Appendix ~\ref{sec:appendix_sup_exp_ana}. \ours consistently demonstrates superior performance by selecting motion plans that align with task descriptions and user preferences, even in scenarios with novel task specifications. In contrast, the baselines either fail to interpret action outcomes effectively, resulting in unsafe behaviors, or experience severe performance degradation in novel task specifications.

\begin{figure}
    \centering
    \includegraphics[width=\linewidth]{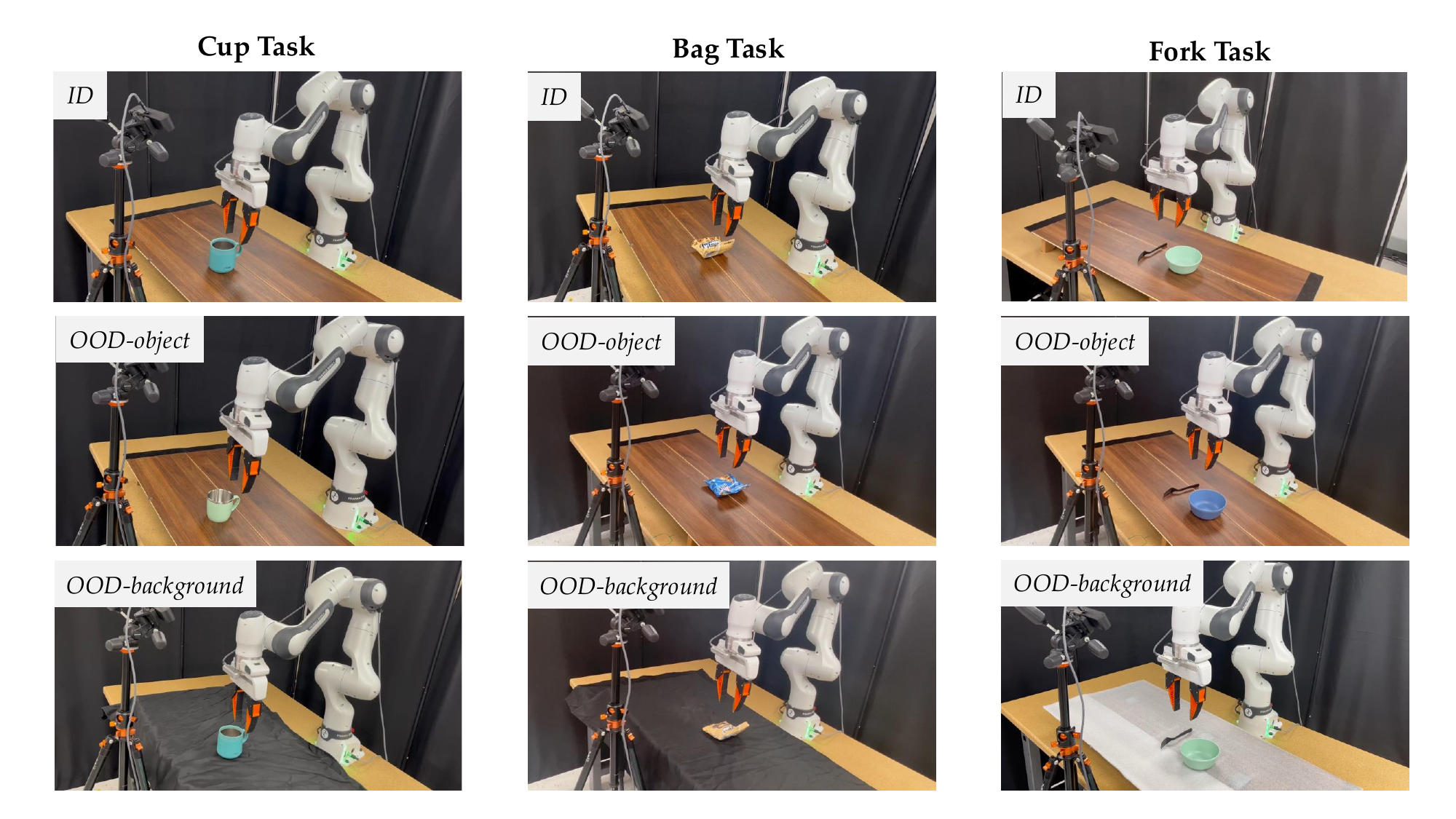}
    \caption{\textbf{Generalization to Environmental Changes.} For each task, we test our method against similar objects of different colors and sizes and also change the table cover to test the system against background variations.}
    \label{fig:ood}
    \vspace{-0.3cm}
\end{figure}

\para{Results: Out-of-distribution Generalization} Table~\ref{tab:policy_steering_all} has shown our system's robustness to novel task descriptions. Moreover, we further test our system's generalization capability by adding variations in the environment. Specifically, we study the variations in object appearances including colors and sizes and variations in background. Among the six tested scenarios shown in Fig~\ref{fig:ood}, our method can generalize to those variations with small performance drop.

\para{Results: Policy Steering Speed}
Our system queries the VLM twice to first generate behavior narrations and then select the best action plan. The overall inference time is 3.7 seconds among which the generation of behavior narrations for 6 candidate action plans takes 1.3 seconds. In comparison, \vlmact takes 22.0 seconds in total for VLM inference and behavior narration runs 19.7 seconds, much slower than \ours, partially due to the usage of token per patch in images rather than a single token per state (image) like \ours. Additionally, our world model and VLM communicate directly in latent space, avoiding image decoding from world model and encoding from VLM, which could further speeds up inference.

\begin{table}[ht]
    \centering
        \renewcommand{\arraystretch}{1.5}
     \setlength{\tabcolsep}{1pt}
    \begin{tabular}{c|c|c|c|c}
    \hline
         \multirow{2}{*}{Method} &  \multicolumn{4}{c}{Time (Seconds) $\downarrow$}\\
         \cline{2-5}
         
         &World Model Prediction & Behavior Narration& Evaluation & Total \\
         \hline
        \ours &  0.1 &  1.3 & \multirow{2}{*}{2.3} & 3.7\\
        \vlmact  & - & 19.7 & & 22.0\\
        \hline
    \end{tabular}
    \caption{\textbf{Inference time for the Policy Steering System.} Inference time for each component in the system (averaged across 3 runs) shows that \ours greatly reduces the time to generate behavior narrations from our modified VLM compared to \vlmact. }
    \label{tab:inference time}
\end{table}

\para{Supplementary Experiments \& Analyses} To provide an in-depth analysis of our approach, we performed additional studies on the impact of each component on the overall system and a detailed comparison with baselines. 
We also showcase an additional application of our system as a runtime monitor which returns if a single action plan is good or bad, opening up potential future directions for soliciting supervisor assistance. 
All these experiments can be found in App.~\ref{sec:appendix_sup_exp_ana}.

\section{Limitations}
While \textbf{FOREWARN} exhibits strong policy steering across diverse task settings, it is not without limitations. First, it assumes base policy is sufficiently competent—i.e., already containing the correct behavior. Future work should investigate how to detect if none of the policy’s generated action plans are suitable for the deployment context, and how to improve the base policy via targeted fine-tuning data.

A second limitation lies in modeling real-world interaction dynamics. Our component-level analysis in App.~\ref{sec:appendix_sup_exp_ana} revealed that our system’s primary failures stem from the world model’s imprecise ``imagination'', exacerbated by our limited training data. More advanced visual features (e.g., DINO features~\citep{oquabdinov2}) might improve the world model’s robustness to visual distractors. Ongoing research on large-scale, generalizable world models~\citep{wu2024ivideogpt} for manipulation may also inform future extension of this framework.

Another bottleneck of the current system is the inference overhead, a common challenge for large autoregressive models like LLMs and world models. 
Some approaches to this include: 1) hierarchical decomposition of \textit{FOREWARN} running  high-level reasoning at low frequency while maintaining low-level control at high frequency like the common practice of hierarchical Vision Language Action Models (VLAs), 2) using advancement in quantization and caching techniques, 3) distilling our latent-aligned VLM into a model with smaller size, 4) strategically allocate inference time by identifying keypoints when it is ``worth'' reasoning for longer or deliberately.  

Finally, our VLM is built upon an open-source LLama-3.2-11B-Vision-Instruct model, whose visual reasoning capabilities lag behind its language-based commonsense reasoning. A stronger backbone could yield higher-quality behavior descriptions, especially if it excels at video captioning tasks. 

\section{Conclusion}
In this work, we investigate the problem of policy steering for multi-modal generative policies. We propose a novel framework \textbf{FOREWARN} that unlocks Vision Language Models (VLMs) to serve as open-vocabulary verifiers for run-time policy steering. 
Our method decouples the steering problem as future prediction (foresight) and outcome assessment (forethought). Through an explicit world model and a latent-space alignment strategy, we enable VLMs to reason about sensorimotor data using natural language. Our experiments across diverse manipulation tasks confirm that \textbf{FOREWARN} not only provides interpretable and reliable failure detection, but also significantly enhances policy success rates through flexible, generalizable steering. These results highlight the promise of combining learned dynamics models with language-based reasoning to improve test-time performance of robot policies.

\section*{Acknowledgments}
This work is funded in part by a Google Research Scholar Award. Gokul Swamy is supported by STTR grant. We also want to thank Junwon Seo for providing feedback for the paper and Michelle Zhao, Pranay Gupta, Kensuke Nakamura, Lasse Peters for the helpful discussions of ideas.

\bibliographystyle{plainnat}
\bibliography{bibliography}
\clearpage
\appendix
\subsection{Algorithm \& Implementation Details}

\subsubsection{Base Policy} \hfill 

\label{sec:appendix_base_policy} We use Diffusion Policy~\cite{chi2024diffusionpolicy} as our robot action generation model due to its strong ability to capture multimodal and complex robot behaviors.
 
 \begin{figure}[ht]
     \centering
     \includegraphics[width=\linewidth]{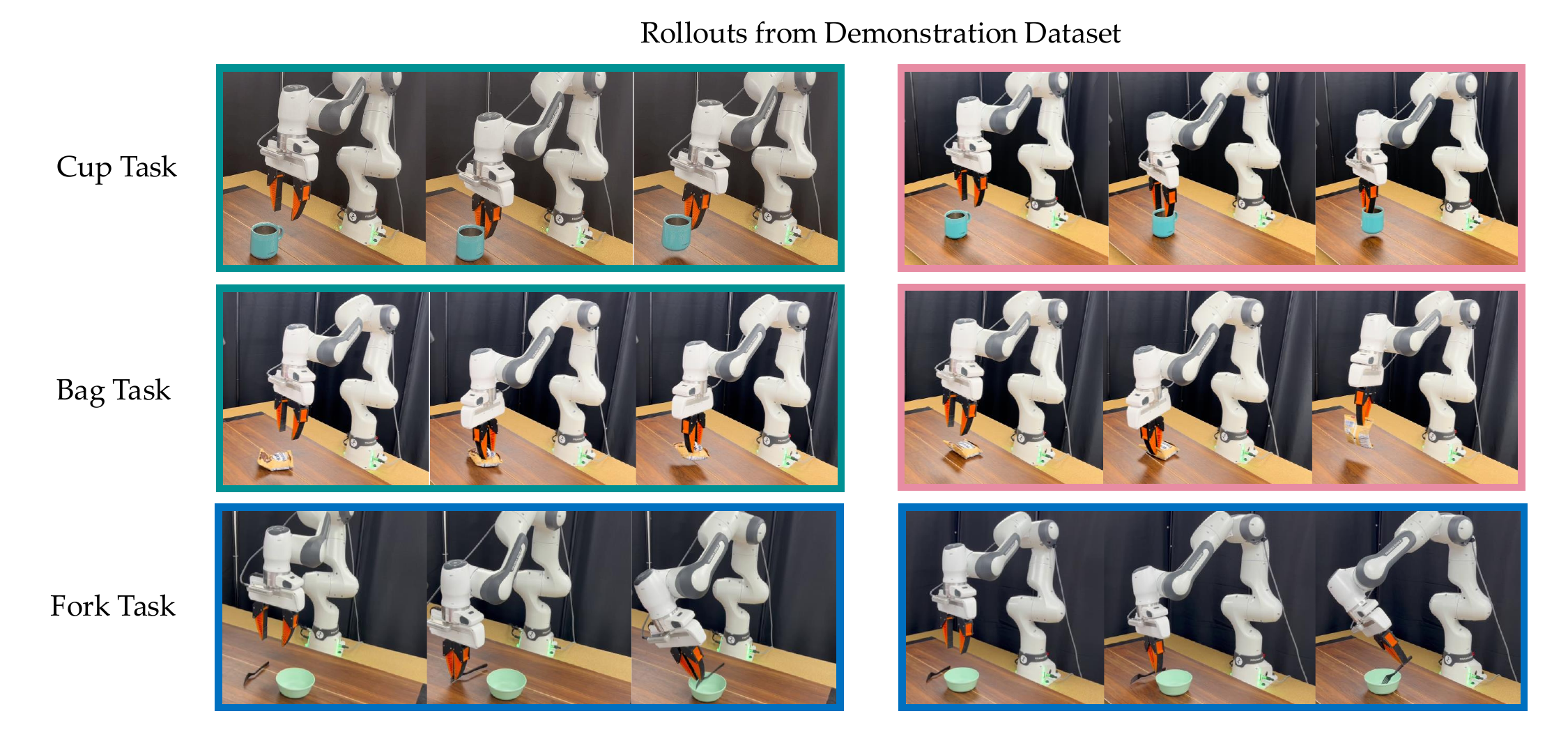}
     \caption{\textbf{Multimodality in Demonstration Datasets.}}
     \label{fig:demonstration_rollout}
 \end{figure}

\para{Training dataset} For each task, we collect 100 multimodal demonstrations and train the policy on an NVIDIA A6000 GPU following the procedure in~\citep{chi2024diffusionpolicy}.  In \textbf{Cup Task},the training dataset consists of 50 demonstrations where the robot grasps the cup by the rim while touching the inner surface, and 50 demonstrations where the robot grasps it by the handle. In \textbf{Bag Task}, the training dataset consists of 50 demonstrations where the robot grasps the middle part of the bag and 50 demonstrations where it grasps the top edge without crushing the chips. 
In \textbf{Fork Task}, the training dataset consists of 4 modes, including 1) picking up the fork by tines and dropping high; 2) picking up the fork by the handle and dropping high; 3) picking up the fork by the tines and dropping low; 4) picking up the fork by the handle and dropping low. Each mode has 25 demonstrations.
Figure~\ref{fig:demonstration_rollout} presents multimodal demonstrations for each task, and Table~\ref{tab:base_policy} details the hyperparameters used for training. 
\begin{table}[ht]
    \centering
    \begin{tabular}{c|c}
        Hyperparameter & Value  \\
        \hline
         State Normalization & Yes \\
         Action Normalization & Yes \\
         Image Chunk & 2  \\
         Image Size & 256 \\
         State Dimension & 8 \\
         Action Dimension & 10 \\
         Action Execution Horizon (Bag) & 140 \\
         Action Execution Horizon (Cup) & 120 \\
          Prediction Horizon (Bag) & 120 \\
         Prediction Horizon (Cup) & 140 \\
         Batch Size & 100 \\
         Training Epochs &  600 \\
         Learning Rate &  1e-5\\
         Number of Worker& 16 \\
         Train Diffusion Step & 100 \\
         Inference Diffusion Iteration & 16 \\
         
         \hline
   
    \end{tabular}
    \caption{\textbf{Base Policy Hyperparameters.}}
    \label{tab:base_policy}
\end{table}

\para{Observation and action spaces} The robot’s policy receives inputs from a wrist-mounted camera, a third-person camera, and proprioceptive states—including the end-effector position, orientation quaternion relative to the base, and gripper opening—to predict actions that control the end-effector’s absolute pose and gripper opening.

\para{Action plan aggregation} During policy steering, we aggregate 100 sampled action plans into 6 modes. We represent each action trajectory as a T $\times$ 7 array, where T is the trajectory length, and each action consists of the gripper’s position (3D coordinates) and orientation (quaternion with four components). To cluster these trajectories into 6 modes, we apply Time Series K-Means with Dynamic Time Warping (DTW) as the distance metric. This allows us to group similar motion patterns while accounting for temporal variations in trajectory execution. The average trajectory within each cluster is used as the aggregated action plan. In the subsequent stages of our system, these 6 aggregated action plans serve as candidate options for policy steering selection.
\\

\subsubsection{World Model}\hfill 

\label{sec:appendix_world_model}
\para{Motivation} The effectiveness of world models has been demonstrated across various embodied domains~\citep{liumulti,wu2023daydreamer}. In our problem setting, it provides several key advantages: 1) it grounds low-level actions, difficult for a VLM to interpret—by predicting future image observations from action plans, bridging the gap between low-level actions and visual observations; 2) it compresses information into latent states that not only retain essential details for high-quality image decoding but also effectively predict the next latent state given an action. 

\para{Architecture} We use DreamerV3, a state-of-the-art recurrent world model from~\citep{hafner2023mastering}. Our world model, denoted as $\mathcal{W}_\phi = (\enc, \dyn, \dec)$, consists of three key components: an encoder network $\enc$, a recurrent dynamics model $\dyn$ that operates over a stochastic continuous latent space, and a decoder network $\dec$ that projects latent embeddings back into observations. Below, we provide a detailed definition of each module:
\begin{align}
\label{eq:encoder}
\latent_\tau &:= 
\begin{cases} 
\enc(\obs_\tau^i), & \quad \tau = t. \\ 
\enc(\obs_\tau^i, \latent_{\tau-1}^i, \action_{\tau-1}^i), & \quad t < \tau < t + T .
\end{cases} \\
\label{eq:dyancmis}
\hat{\latent}_{\tau+1}^i &\sim
\begin{cases} 
\dyn(\latent_\tau^i, \action_\tau^i), & \quad \tau = t. \\ 
\dyn(\hat{\latent}_\tau^i, \action_\tau^i), & \quad t < \tau < t + T. 
\end{cases} \\
\label{eq:decoder}
\hat{o}^i_\tau &:= \dec(\hat{\latent}_\tau^i), \quad t \leq \tau < t + T .
\end{align}

\begin{table}[ht]
    \centering
    \begin{tabular}{c|c}
        Hyperparameter & Value  \\
        \hline
         Observation Normalization & Yes \\
         Action Normalization & Yes \\
         Image Size & (64, 64) \\
         Batch Length & 64 \\
         Batch Size & 16\\
         Training Step (Cup) & 100000\\
        Training Step (Bag) & 150000\\
        Hidden State $h$ Dimension & 512 \\
        Stochastic Representation $z$ Dimension & 32 \\
        Dynamics Loss Ratio $\alpha_{dyn}$& 0.5 \\
        Representation Loss Ratio $\alpha_{rep}$ & 0.1 \\
        Reconstruction Loss Ratio $\alpha_{pred}$ & 1.0 \\
        CNN Encoder Depth & 32\\
        CNN Encoder Kernel Size & 4\\
        \hline
        
    \end{tabular}
    \caption{\textbf{World Model Hyperparameters.}}
    \label{tab:world_model}
\end{table}
\para{Training} The world model $\mathcal{W}_\phi$ is pretrained using an offline dataset of policy's rollouts and demonstrations $\mathbb{D}_{\text{WM}} = \{\{ (\obs_\tau^j, \action_\tau^j, \obs_{\tau+1}^j)\}_{\tau=0}^{H-1}\}_{j=1}^M$. Apart from 100 demonstrations, we collect 250 rollouts for each task, including both the success and failure experiences of the generative policy $\pi$ deployed on the real robot. 
\begin{figure}[ht]
    \centering
    \includegraphics[width=\linewidth]{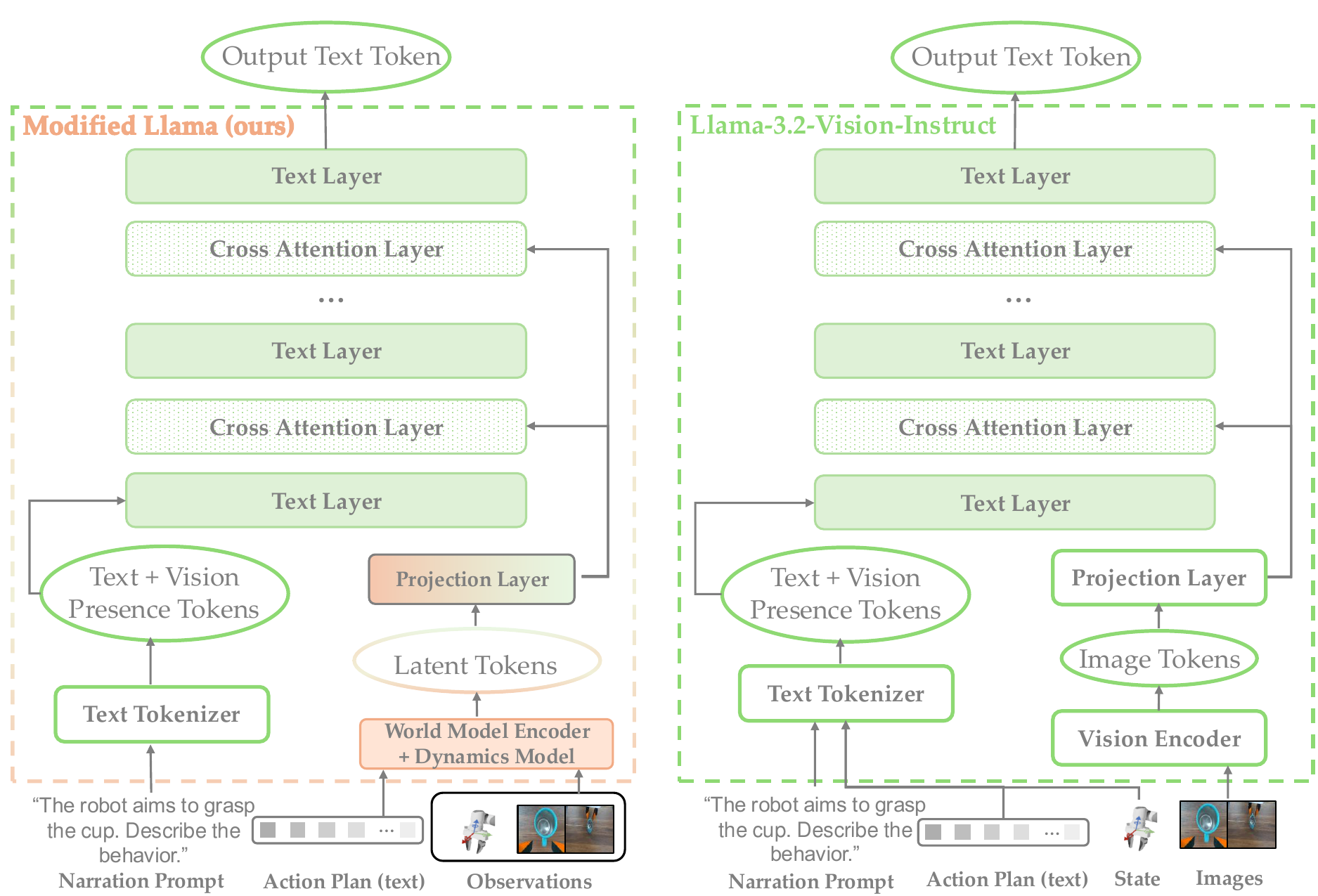}
    \caption{\textbf{VLM Architecture}. On the left is detailed architecture \ours, an modified VLM with an explicit world model. On the right is original Llama-3.2-Vision-Instruct model, which is used as our baseline \vlmact. This is an elaborated version of Fig.~\ref{fig:overview_fig}}
    \label{fig:llama_structure}
\end{figure}

Since offline datasets lack reward signals, we omit the reward loss. Instead, the pretraining of the world model is supervised using a modified loss function:  $\mathcal{L}_\mathcal{W} = \alpha_{dyn} \times \mathcal{L}_{dyn} + \alpha_{rep} \times \mathcal{L}_{rep} + \alpha_{pred} \times \mathcal{L}_{pred} $. Here, the dynamic loss $ \mathcal{L}_{dyn}$ incentivizes accurate forward predictions in the latent space while $\mathcal{L}_{rep}$ ensures that latent states are informative for reconstructing observations and learning good representation. Finally, the prediction loss $ \mathcal{L}_{pred}$ minimizes the error between decoded observations of the predicted latent states and the ground-truth observations. The exact loss calculation is shown in ~\citep{hafner2023mastering}. The hyperparameters used for training world model is shown in Table~\ref{tab:world_model}.

During training, the decoder $\dec$ reconstructs observations from the latent space to compute the prediction loss and the visualization helps us select the best world model. 
\begin{table}[ht]
    \centering
  \setlength{\tabcolsep}{5pt}
    \renewcommand{\arraystretch}{1.5}
    \begin{tabular}{c|c}
        Hyperparameter & Value  \\
        \hline
        LoRA (rank) & 8 \\
        LoRA (dropout) & 0.05  \\
        LoRA (alpha) & 32 \\
        Precision & bfloat16 \\
        Batch Size & 10 \\
        Learning Rate & 1e-4 \\
        Epoch (Cup) & 10 \\
        Epoch (Bag) & 15 \\
        \multirow{2}{*}{Finetuned Layers} & ["down\textunderscore proj", "up\textunderscore proj", "o\textunderscore proj", "k\textunderscore proj", \\
        & "q\textunderscore proj",  "v\textunderscore proj", "gate\textunderscore proj", "linear\textunderscore proj"]\\
        \hline
    \end{tabular}
    \caption{\textbf{VLM Hyperparameters of \ours}}
    \label{tab:vlm_hyperparameter}
\end{table}

\para{Deployment} 
Because we cannot reset the real-world environment to the exact same state, each action sequence $\acttraj_t^i$
  yields only a single observation sample $\obstraj^i_t$. As a result, the expectation in Eq.~\ref{eq:vlm_reward} is approximated in practice. During finetuning, we make the dynamics model $f_\phi$ deterministic by taking the mean of its predicted distribution. Although the model provides a 64-step prediction horizon, we downsample these future latent states to 16 to reduce redundancy from minimal changes across adjacent steps. Each latent state is formed by concatenating the hidden state and the stochastic representation, and we keep the world model frozen throughout VLM finetuning and deployment.
\\
\begin{figure}[h!]
    \centering
    \includegraphics[width=\linewidth]{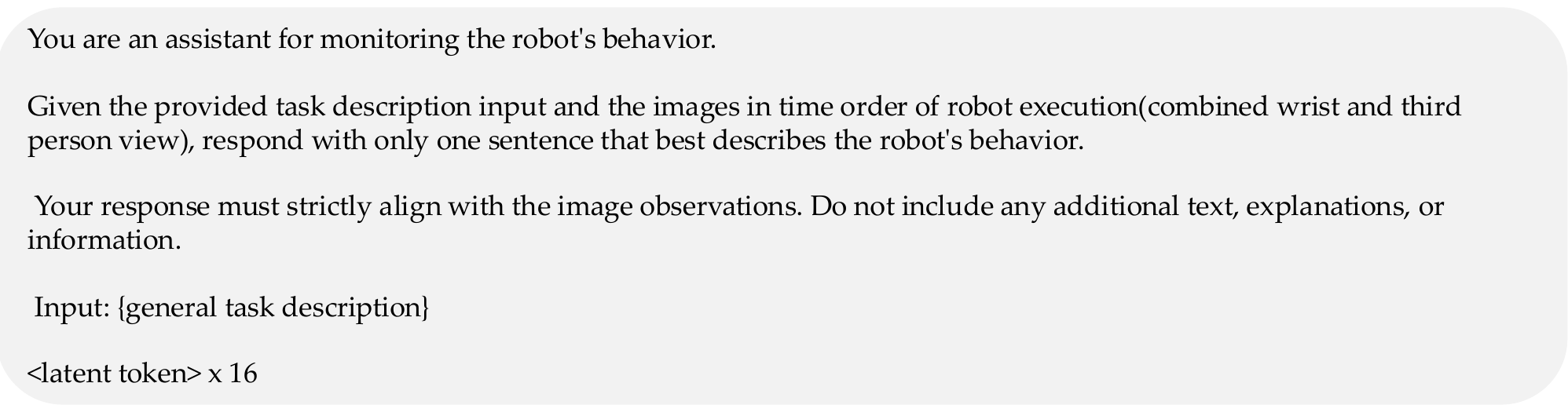}
    \caption{\textbf{Prompt Template for Behavior Narration in \ours.}}
    \label{fig:prompt_behavior}
\end{figure}

\subsubsection{Vision Language Model}\hfill

\label{sec:appendix_vlm}

\para{Architecture} We use Llama-3.2-11B-Vision-Instruct model as our VLM backbone. We modify the original Llama Model to incorporate the explicit world model to predict outcomes of the action plans first and then use VLM to reason about the latent states to generate behavior narrations. The modified architecture is visualized in Fig.~\ref{fig:llama_structure}.
Specifically, we replace the original vision encoder (ViT) of Llama with our world model's encoder as well as dynamics model,  and project latent states as text tokens. We use one single linear layer to project latent tokens to text tokens.

\begin{figure}[h!]
    \centering
    \includegraphics[width=\linewidth]{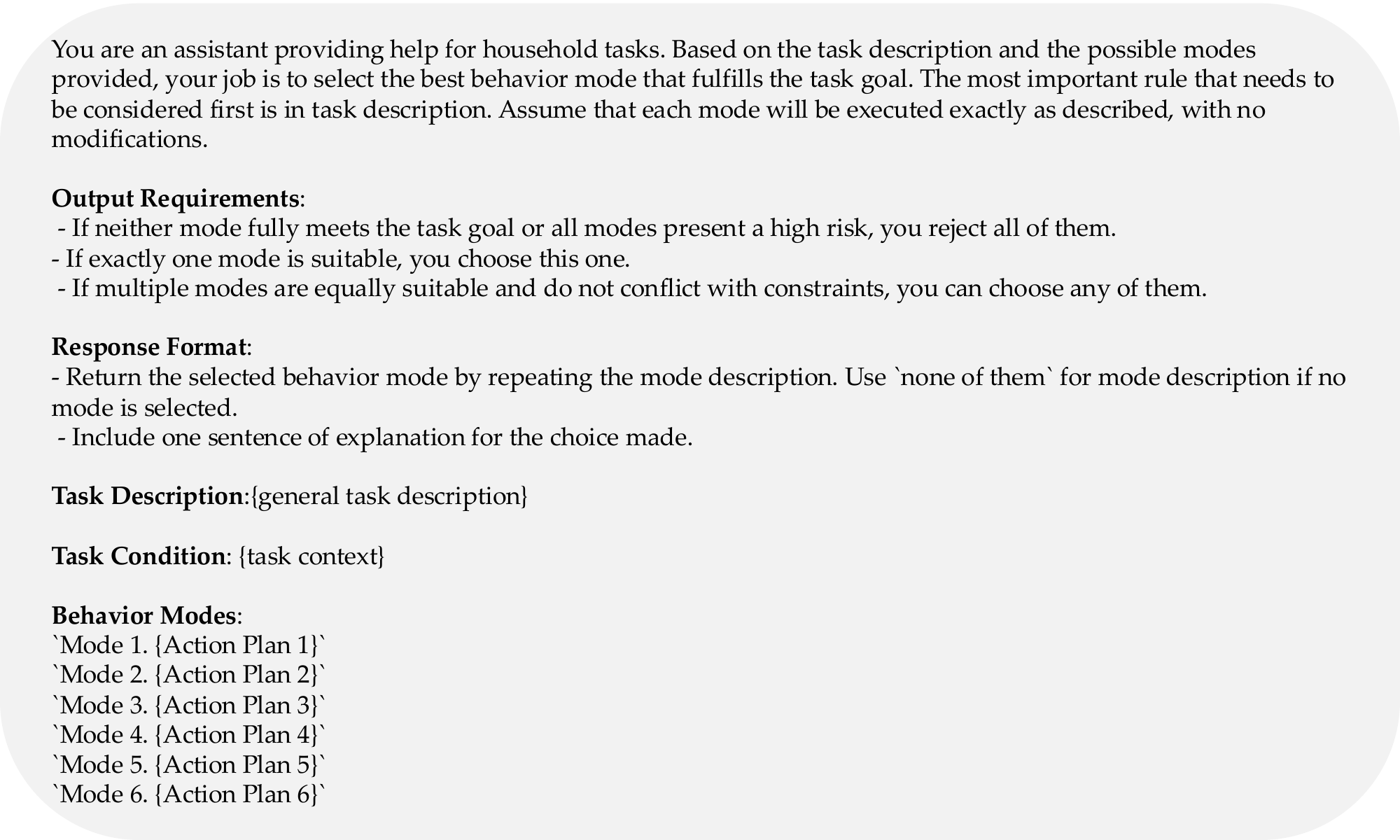}
    \caption{\textbf{Prompt Template for Policy Steering.}}
    \label{fig:prompt_steering}
\end{figure}

\para{Finetuning} We adopt LoRA to finetune our modified model, loading the base weights from Llama-3.2-Vision-Instruct and randomly initializing the new linear projection layer. This approach updates only $0.2664\%$ of the original model’s parameters. The finetuning hyperparameters are listed in Table~\ref{tab:vlm_hyperparameter}, and we select the model with the lowest evaluation loss for deployment.
\begin{figure}[h!]
    \centering
    \includegraphics[width=\linewidth]{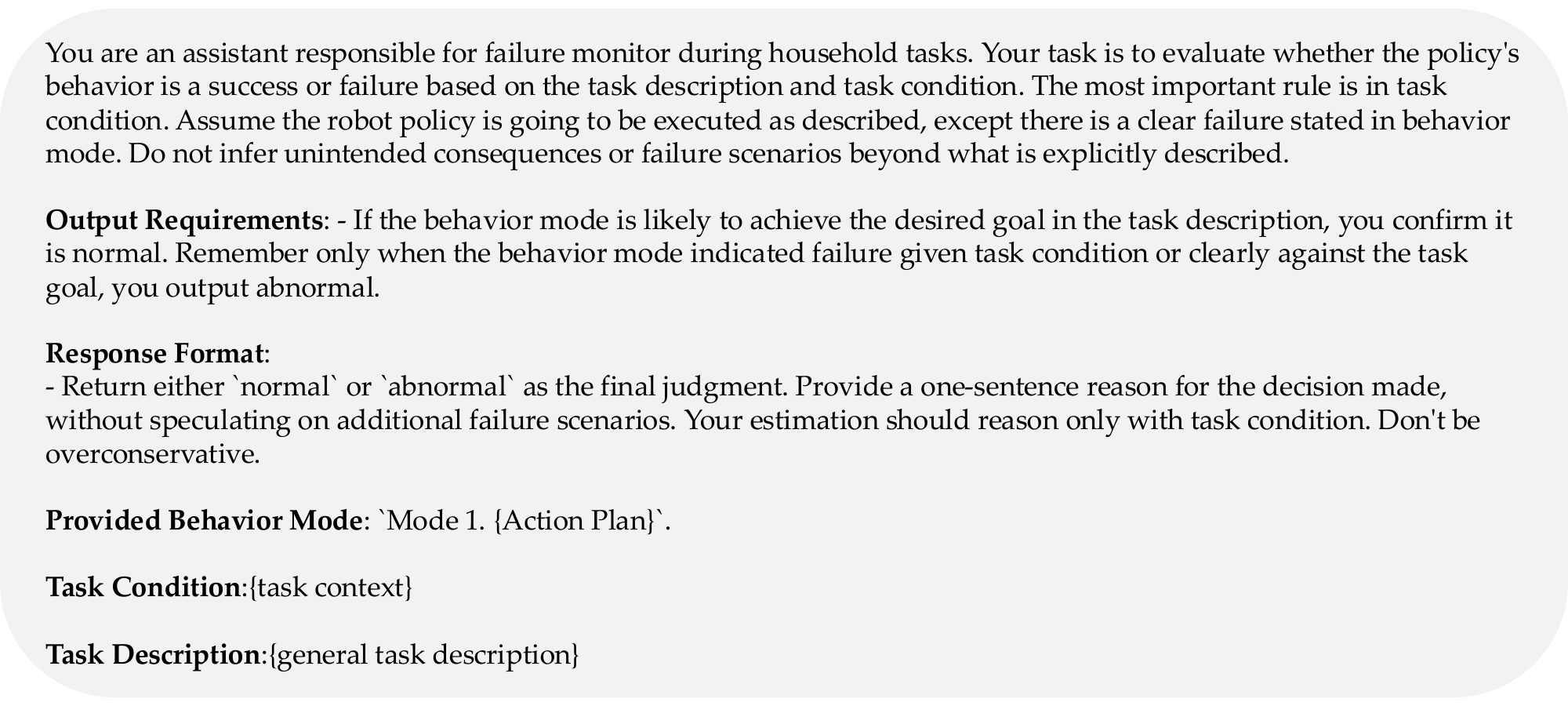}
    \caption{\textbf{Prompt Template for Failure Monitoring.}}
    \label{fig:prompt_monitoring}
\end{figure}

\para{Prompt}
The modified VLM is finetuned to generate behavior narration with the prompt template in Fig.~\ref{fig:prompt_behavior}, explaining the predicted latent states from the world model. 

In policy steering. we sample and aggregate action sequences into 6 modes to query VLM for the best choice. Each mode's behavior narration replaces Action Plan in the prompt template in Fig.~\ref{fig:prompt_steering}. 

Another potential usage of our system is to evaluate and monitor different policies' performances under specific task description before execution. To showcase our system can be a reliable failure monitor across different tasks, we conduct additional experiments in App.~\ref{sec:appendix_sup_exp_ana}. We list the prompts used for failure monitor in Fig.~\ref{fig:prompt_monitoring}.

Different tasks have different task descriptions, and specifications (i.e., contexts), which are put in the prompt for policy monitoring and policy steering can be found in Sec.~\ref{sec:experiments}
 
\begin{figure}[h!]
        \centering
        \includegraphics[width=\linewidth]{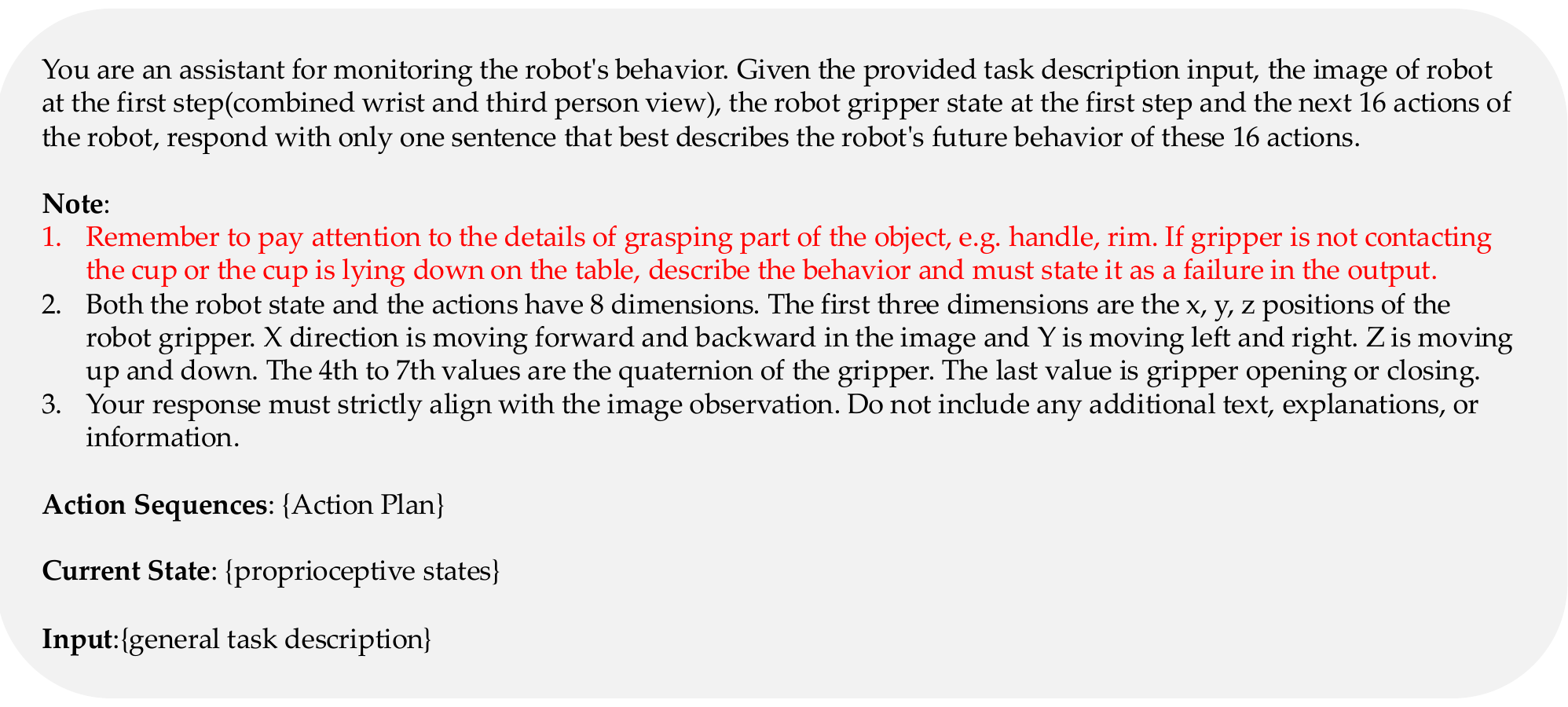}
        \caption{\textbf{Prompt Template for Behavior Narration for \vlmact in Cup Task}}
        \label{fig:prompt_vlmact}
    \end{figure}
\begin{table}[h!]
    \centering 
    \begin{tabular}{c|c}
   Hyperparameter & Value  \\
        \hline
        LoRA (rank) & 8 \\
        LoRA (dropout) & 0.05  \\
        LoRA (alpha) & 32 \\
        Quantization & 4bit \\
        Batch Size & 1 \\
        Learning Rate & 1e-5 \\
        Epoch (Cup) & 10 \\
        Epoch (Bag) & 15 \\
        \multirow{2}{*}{Finetuned Layers} & ["down\textunderscore proj", "up\textunderscore proj", "o\textunderscore proj", "k\textunderscore proj", \\
        & "q\textunderscore proj",  "v\textunderscore proj", "gate\textunderscore proj"]\\
        \hline
    \end{tabular}
    \caption{VLM Hyperparameters of \vlmact.}
    \label{tab:vlmact_hyperparameter}
    \end{table}

\subsubsection{Additional Details of Baselines}\hfill
\label{sec:appendix_additional_baselines}

    \para{\vlmact} This baseline is an ablated version of \ours without the explicit world model. It uses the original Llama-3.2-11B-Vision-Instruct model as shown in right part of Fig.~\ref{fig:llama_structure} and finetuned with the same labels as in \textbf{VQA Dataset}. The action plans and states are prompted as text and image observations are concatenated together and processed as image tokens. Details of the prompt are available in Fig.~\ref{fig:prompt_vlmact}. Since VLM cannot directly interpret the low-level action control, we give some privileged information in the prompt, marked as red in the prompt template, to help it generate behavior narration. However, this baseline still struggles to generate accurate behavior narration as shown in Sec.~\ref{sec:behavior}, despite being finetuned on the same dataset. For policy monitoring and policy steering, \vlmact uses the same prompt as \ours. Hyperparamers used for finetuning are shown in Table.~\ref{tab:vlmact_hyperparameter}

    \begin{figure}[h!]
        \centering
        \includegraphics[width=\linewidth]{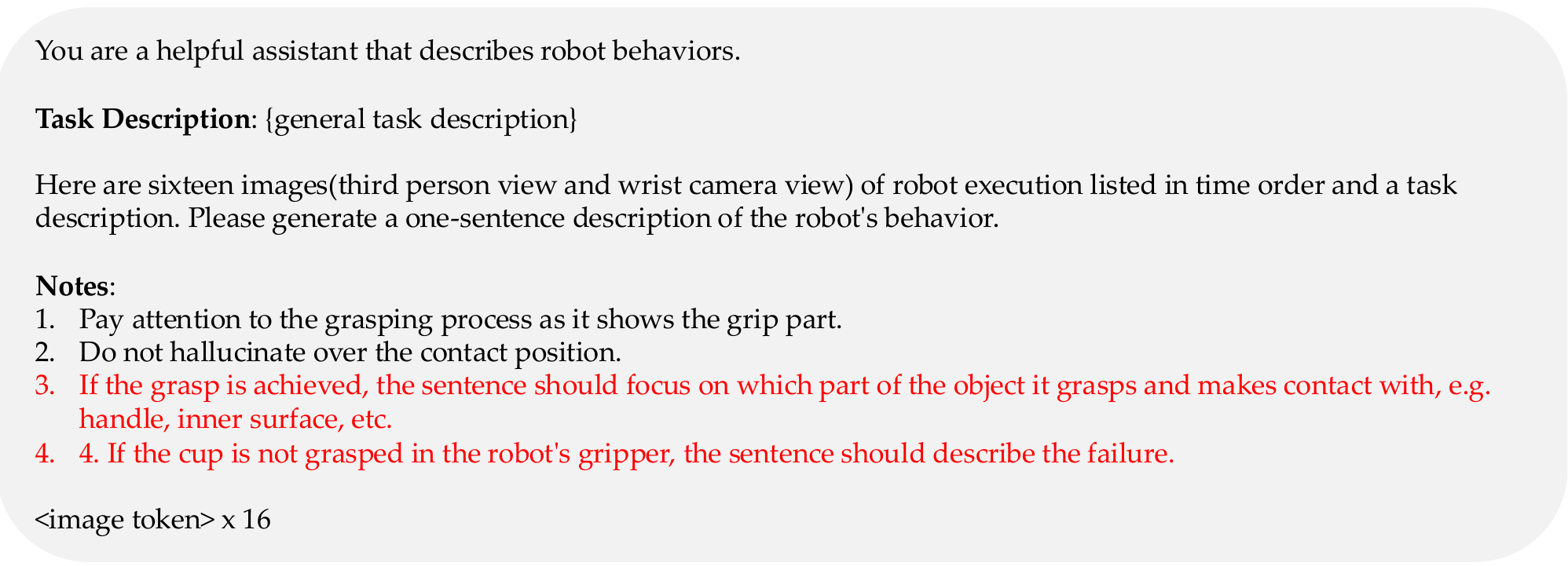}
        \caption{\textbf{Prompt Template for Behavior Narration for GPT-4o in Cup Task.}}
        \label{fig:prompt_gpt4o}
    \end{figure}
       \begin{table}[h!]
    \centering 
    \begin{tabular}{c|c}
   Hyperparameter & Value  \\
        \hline
        Embedding Dimension & 64 \\
        Number of Head & 1 \\
        Attention Dropout & 0.05\\
        Embedding Dropout & 0.05 \\
        Block Output Dropout & 0.05 \\
        Context Length & 16 \\
        Sinusoidal Embedding & True \\
        Learning Rate & 1e-4\\
        Gradient Clip & $(-\infty , 100]$ \\
        Epochs & 30 \\
        
        \hline
    \end{tabular}
    \caption{Hyperparameters of \classdynlatent.}
    \label{tab:classifier}
    \end{table}
    \para{\vlmimg \& \vlmimgoracle} We further evaluate an advanced VLM (GPT-4o) to interpret fine-grained motion details from a sequence of 16 images, either reconstructed from predicted latent states or recorded from actual execution. GPT-4o runs at the default temperature. As shown by the red text in Fig.~\ref{fig:prompt_gpt4o}, GPT-4o still struggles to comprehend subtle motion details, even when given privileged information to guide its focus.

    \para{\classdynlatent} For policy steering and monitoring, we adopt a causal transformer-based binary classifier~\citep{liumulti} to directly predict success or failure from future latent states. Table~\ref{tab:classifier} summarizes its hyperparameters. We manually label each rollout as success or failure under the first task description to train the classifier.
    
\begin{figure}[h!]
        \centering
        \includegraphics[width=\linewidth]{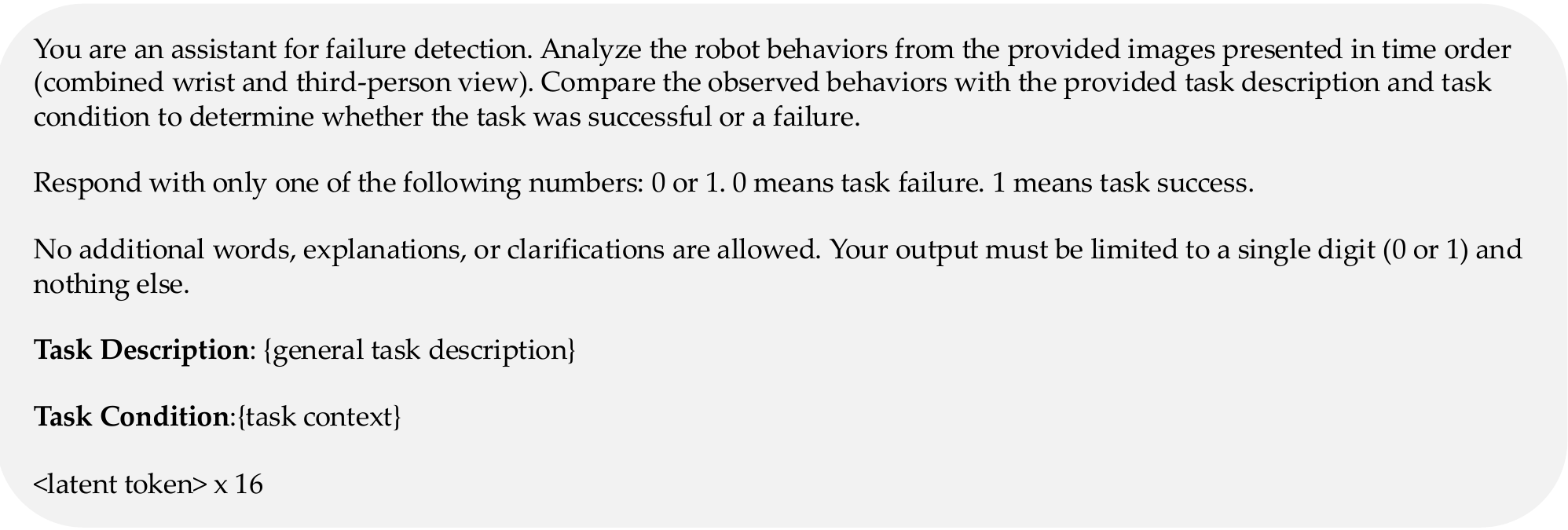}
        \caption{\textbf{Prompt Template for Policy Monitoring for \vlmdynlatentbin.}}
        \label{fig:prompt_vlmdynlatentbin}
    \end{figure}

    \para{\vlmdynlatentbin}  This end-to-end baseline also predicts a binary success/failure label from future latent states, but employs the same modified VLM architecture as \ours, which provides a larger capacity than the transformer-based classifier. It is trained on the same dataset as \classdynlatent, with the prompt template shown in Fig.~\ref{fig:prompt_vlmdynlatentbin}.

            \begin{figure}[h!]
        \centering
        \includegraphics[width=\linewidth]{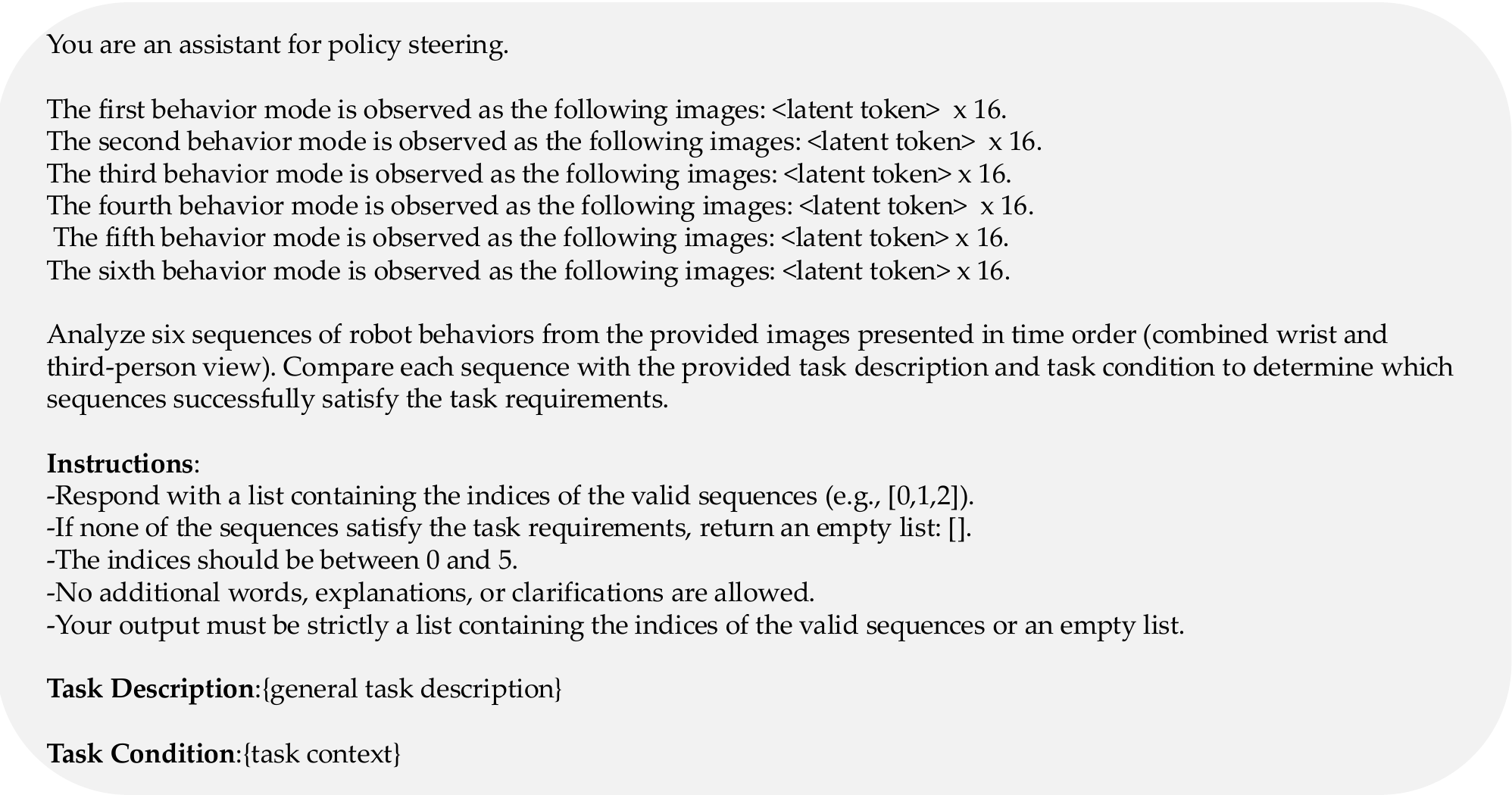}
        \caption{\textbf{Prompt Template for Policy Steering for \vlmdynlatentcat.}}
        \label{fig:prompt_vlmdynlatentcat}
    \end{figure}
    \para{\vlmdynlatentcat} 
Similarly, we develop an end-to-end approach that directly predicts a valid set of action plans for policy steering. The modified VLM is finetuned to output which action‐plan indexes are valid under the current task description. Its input includes a text prompt and six sequences of predicted future latent states (Fig.~\ref{fig:prompt_vlmdynlatentcat}), each corresponding to one candidate action plan.

\subsection{Experiment}
\subsubsection{Real Robot Setup}
\begin{figure}[h!]
    \centering
    \includegraphics[width=\linewidth]{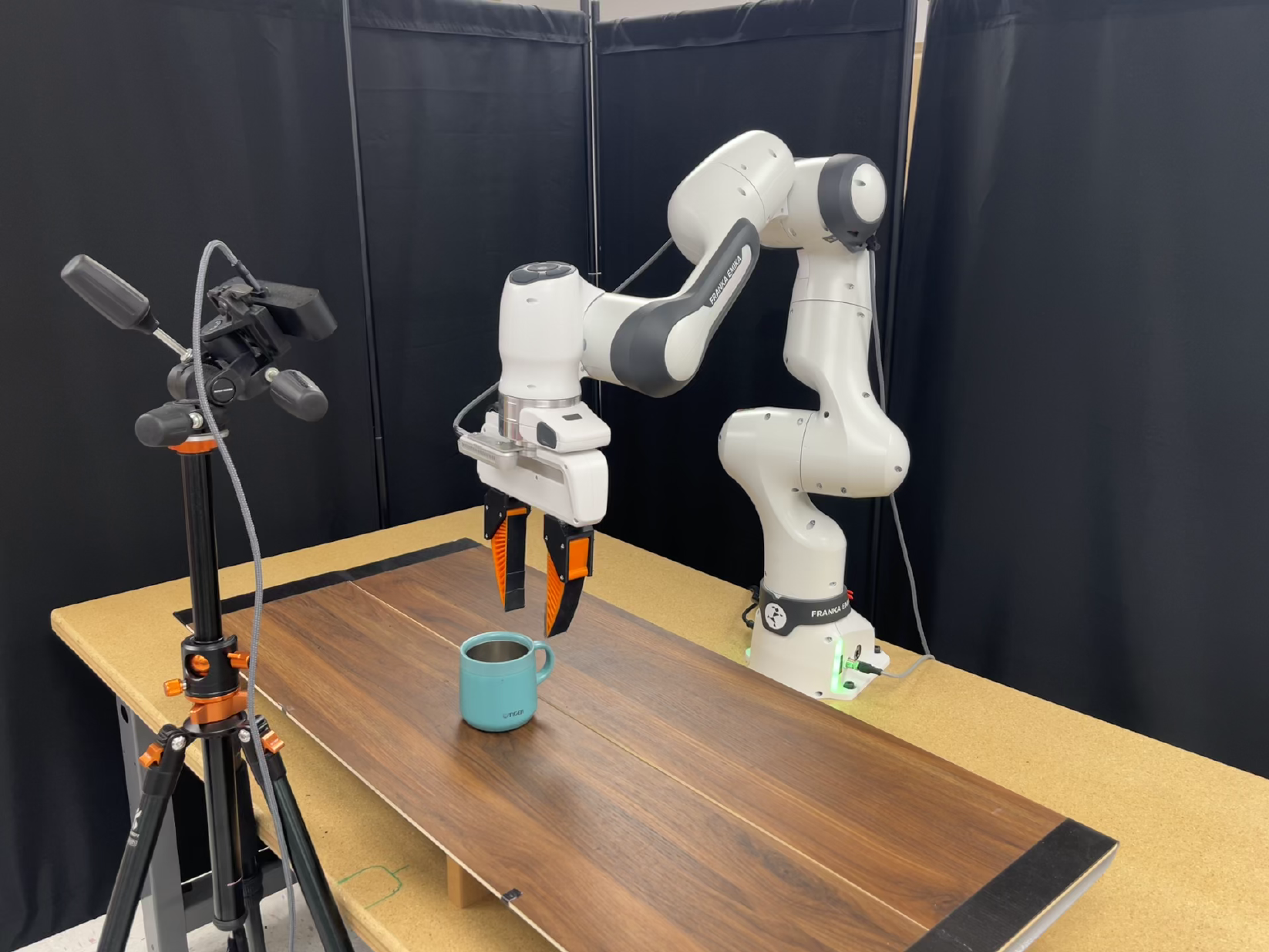}
    \caption{\textbf{Real Robot Setup.} The visualization of the real robot environment and the positions of the cameras. }
    \label{fig:robot_setup}
\end{figure}

\hfill 
\label{sec:appendix_robot_setup}
Fig.~\ref{fig:robot_setup} demonstrates the setup of our real-world experiments. We employ two cameras, a RealSense D435 camera on the Franka hand and a Zed mini 2i camera placed in front of the robot. In order to increase the contact region and compilancy, we replace the original Franka gripper finger with 3D printed gripper finger from~\citep{chi2024universal}.

    \begin{figure}[h!]
    \centering
    \includegraphics[width=\linewidth]{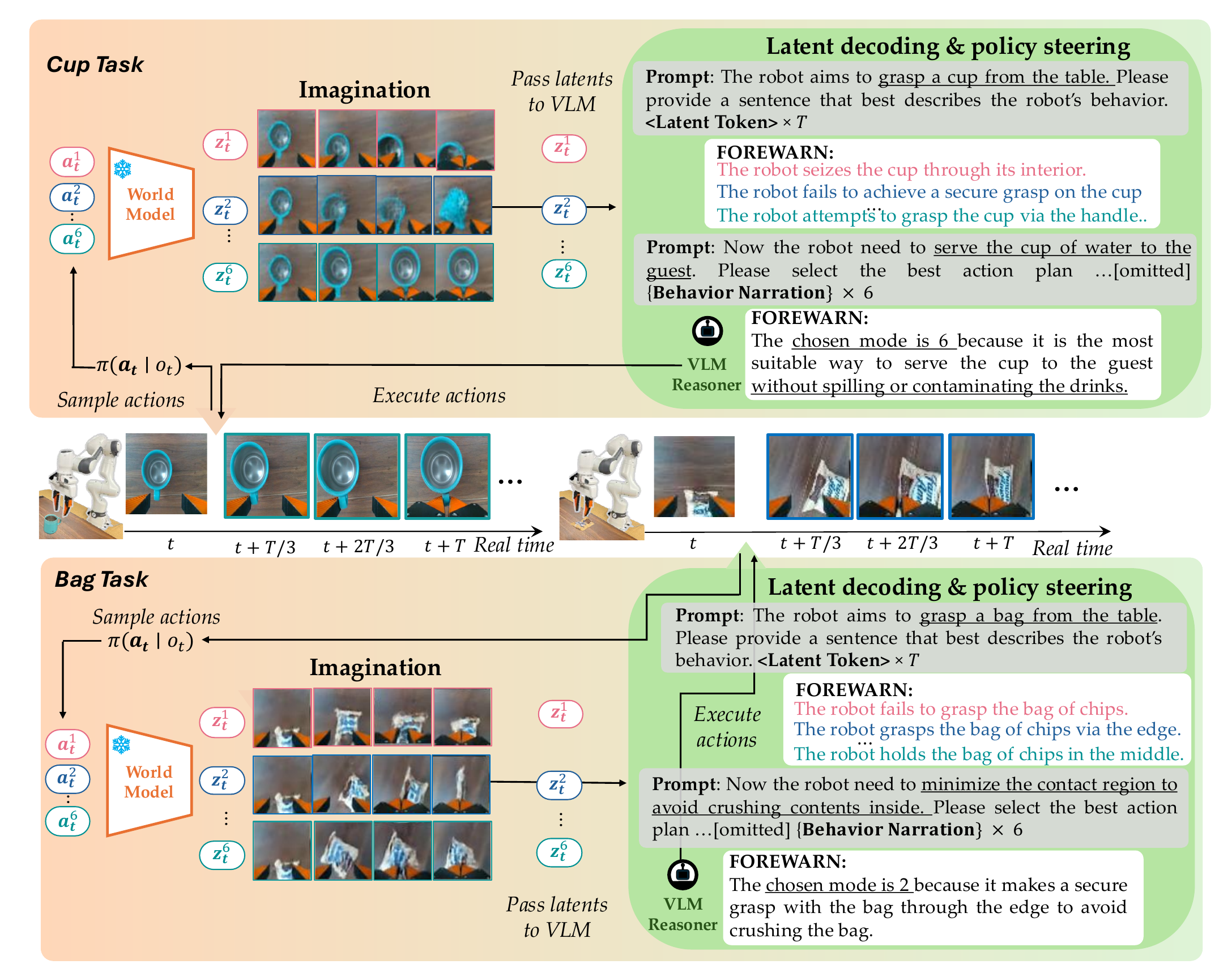}
    \caption{\textbf{Policy Steering for Cup and Bag Task.} We visualize the steering process for the \textbf{Cup} task on the top and the \textbf{Bag} task on the bottom.  For each task, we visualize the imagined $\thor$-step rollouts decoded from the world model for the 3 out of 6 action plans sampled from the base policy on the left. On the right, we show the behavior narrations generated from our finetuned VLM $\mathcal{T}^{\vlm}_\psi$ and the VLM's reasoning $\rewvlm_\psi(\cdot; \lang)$ about the outcomes based on the task description $\lang$ and behavior narrations to select the best action plan to execute. The time axis shows the real-world execution of the selected behavior from the perspective of the wrist-camera.}
    \vspace{-0.3cm}
    \label{fig:policy_steering_cup_bag}
\end{figure}
\begin{table*}[ht]
    \centering
    \setlength{\tabcolsep}{2pt}
    \renewcommand{\arraystretch}{1.5}
    \begin{tabular}{c|ccc|ccc|ccc|ccc|ccc|ccc}
    \hline
    \multirow{3}{*}{Method} 
    & \multicolumn{6}{c|}{Cup} & \multicolumn{6}{c|}{Bag} & \multicolumn{6}{c}{Average} \\
    \cline{2-19}
    & \multicolumn{3}{c|}{Training} & \multicolumn{3}{c|}{Unseen} & \multicolumn{3}{c|}{Training} & \multicolumn{3}{c|}{Unseen} & \multicolumn{3}{c|}{Training} & \multicolumn{3}{c}{Unseen} \\
    \cline{2-19}
    & Acc$\uparrow$& TPR$\uparrow$& TNR$\uparrow$& Acc$\uparrow$& TPR$\uparrow$ & TNR$\uparrow$ & Acc$\uparrow$ & TPR$\uparrow$ & TNR$\uparrow$ & Acc$\uparrow$ & TPR$\uparrow$ & TNR$\uparrow$ & Acc$\uparrow$ & TPR$\uparrow$ & TNR$\uparrow$ & Acc$\uparrow$ & TPR$\uparrow$  & TNR$\uparrow$ \\
    \hline
    \ours (Ours)             & \textbf{0.90} & \textbf{0.80} & \textbf{1.00} & \textbf{0.75} & \textbf{0.70} & 0.80  & 0.75 & 0.71 & 0.77 & \textbf{0.75} & \textbf{0.75} & 0.75 & 0.83 & 0.76 & 0.89 & \textbf{0.75} & \textbf{0.73} & 0.78 \\
    \vlmdynlatentbin   & 0.85 & 0.77 & 0.91 & 0.15 & 0.11 & 0.27 & 0.75 & \textbf{0.86} & 0.69 & 0.40 & 0.38 & 0.42 & 0.80 & \textbf{0.82} & 0.80 & 0.28 & 0.25 & 0.35 \\
    \classdynlatent    & \textbf{0.90} & \textbf{0.80} & \textbf{1.00} & 0.10 & 0.10 & 0.10 & \textbf{0.80} & 0.71 & 0.85 & 0.35 & 0.13 & 0.50 & \textbf{0.85} & 0.76 & \textbf{0.93} & 0.23 & 0.12 & 0.30 \\
    \vlmact            & 0.65 & 0.75 & 0.58 & 0.50 & 0.10 & \textbf{0.90} & 0.60 & 0.00 & \textbf{0.92} & 0.65 & 0.13 & \textbf{1.00} & 0.63 & 0.38 & 0.75 & 0.58 & 0.12 & \textbf{0.95} \\
    \hline
    \end{tabular}
    \caption{\textbf{Policy Monitoring.}
    The reported result in the table is averaged over 20 trajectories. \ours, \vlmdynlatentbin and \classdynlatent perform similarly well in training task description while \vlmact has poor performance. In unseen task description, \ours is the only method that maintains similar performance as training task description.}
    \label{tab:policy_evaluation_all}
\end{table*}

   \begin{figure*}[h!]
       \centering
       \includegraphics[width=\linewidth]{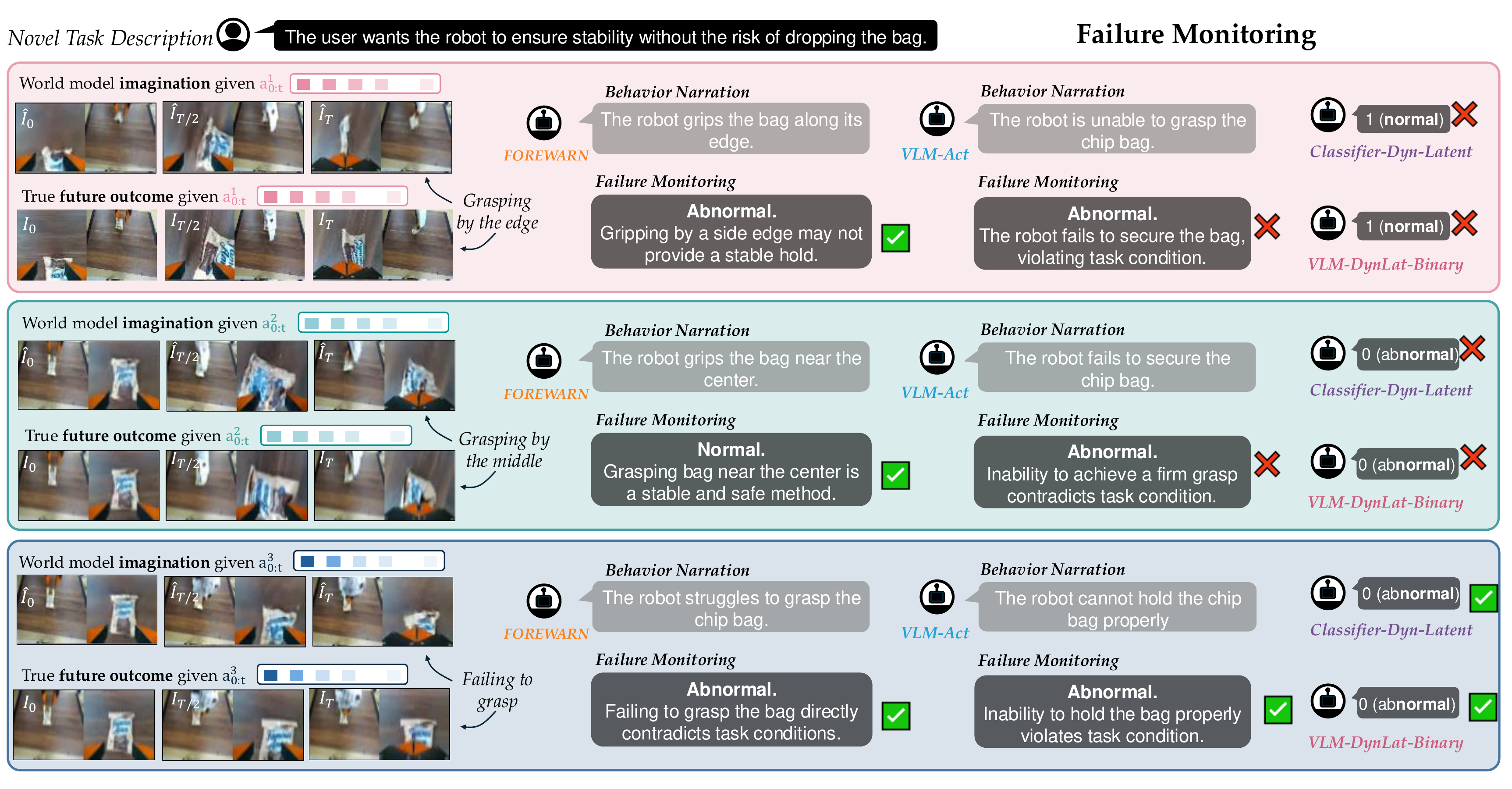}
       \caption{\textbf{Policy Monitoring.} In the top (pink) and medium (green) row, the robot imagines correctly about the robot behaviors but only \ours describes the behavior correctly and generates correct monitoring results with adequate explanations. In the bottom row (blue), all of the methods distinguish failure correctly from success.}
       \label{fig:policy_monitoring_more}
   \end{figure*}

\begin{table*}[h!]
        \centering
        \setlength{\tabcolsep}{3pt}
        \renewcommand{\arraystretch}{1.5}
        
        \begin{tabular}{c|ccc | ccc |ccc |ccc|ccc | ccc}
        \hline
            \multirow{3}{*}{Method} & \multicolumn{18}{c} {Accuracy $\uparrow$}\\
            \cline{2-19}
             & \multicolumn{6}{c|}{Independent}&  \multicolumn{6}{c|}{Training Task Description } & \multicolumn{6}{c}{Novel Task Description } \\
             \cline{2-19}
            & \multicolumn{3}{c}{World Model}& \multicolumn{3}{c|}{Behavior Narration} & \multicolumn{3}{c|}{Reasoning} & \multicolumn{3}{c|}{Overall System} & \multicolumn{3}{c|}{Reasoning} & \multicolumn{3}{c}{Overall System }\\
            \cline{2-19}
            & Cup & Bag & Average & Cup & Bag & Average &Cup & Bag & Average &Cup & Bag & Average &Cup & Bag & Average &Cup & Bag & Average \\
            \hline 
            \ours & \multirow{3}{*}{0.80} &\multirow{3}{*}{0.75}  & \multirow{3}{*}{0.78} &  \textbf{0.90}  & \textbf{0.70} & \textbf{0.80} & \textbf{1.00} & 0.85 & 0.93 & \textbf{0.90} & 0.75 & 0.83  & \textbf{0.90} & 0.85 & 0.88 & \textbf{0.75} & \textbf{0.75} & \textbf{0.75}\\
            \vlmdynlatentbin & &  &   & -& -& - & 0.95 & 0.80 & 0.88 & 0.85 & 0.75 & 0.80 & 0.20 & 0.45 & 0.33  & 0.15 & 0.40 & 0.28 \\
            \classdynlatent &  &  &  & -& -& - & \textbf{1.00}  & 0.90 & \textbf{0.95} & \textbf{0.90} &\textbf{ 0.80 }& \textbf{0.85} & 0.05 & 0.25 & 0.15 & 0.10 & 0.35 & 0.23\\
            \vlmact &- & - & -&0.35 & 0.35 & 0.35 & 0.95 & \textbf{0.95} & \textbf{0.95} & 0.65 & 0.60 & 0.63 & \textbf{0.90}  & \textbf{1.00} & \textbf{0.95} &0.50 & 0.65 & 0.58\\
            \hline
        \end{tabular}
        \caption{\textbf{Performance Breakdown} for each component in our pipeline across four methods. \ours have high accuracy across all components while \vlmact has very low accuracy in narration, leading the poor performance of the overall system. both \classdynlatent and \vlmdynlatentbin performs well in training task description and drop sharply in novel task scenarios. }
        \label{tab:policy_breakdown_all}
    \end{table*}

\subsubsection{Supplementary Experiments \& Analyses}\hfill 
\label{sec:appendix_sup_exp_ana}

\para{More Qualitative Examples for Policy Steering} We include additional qualitative examples for \textbf{Cup} and \textbf{Bag} tasks in Fig~\ref{fig:policy_steering_cup_bag}. These examples further demonstrate the effectiveness of our policy steering system for different tasks. The imagined image sequences from the world model show that our system is capable to predict the various outcomes for different action sequences and the VLM can reason about behavior narrations and correctly evaluate the action plans.

\para{FOREWARN as a VLM-in-the-loop Failure Monitoring System} In this section, we present another application of our approach—preemptive failure monitoring—based on the behavior narrations generated in Sec.~\ref{sec:behavior}. Similarly, our modified VLM first decodes the predicted latent states from the world model, as behavior narrations $\hat{\behavior}$. Then it reasons about behavior narrations under task description $\lang$ and decides whether the future action plan, translated as $\hat{\behavior}$, is a failure. As both quantitative and qualitative results demonstrate,  \ours is a reliable and versatile failure monitoring framework across diverse task.

\para{Baselines} We consider three methods as our baselines, which preemptively predict the outcome of the action plan before the execution. 1) \vlmdynlatentbin takes the predicted latent states from the world model and task description $\lang$ as input to the VLM, directly generating binary output to indicate success or failure without the intermediate step of behavior narration; 2) \classdynlatent takes the predicted latent states as input and trains a transformer-based binary classifier to generate binary output with the same dataset as \vlmdynlatentbin; 3) \vlmact uses generated behavior narrations in Sec.~\ref{sec:behavior} and queries the VLM again to decide if the behavior is a success or failure within the context of the task description $\lang$. These baselines are equivalent to those in Sec.~\ref{sec:steering}.

\para{Metrics}
To evaluate the overall performance of different methods as preemptive failure monitors, we report the standard detection metrics from prior work~\citep{agiaunpacking} including \textit{Accuracy (ACC)}, \textit{True Positive Rate (TPR)}, \textit{True Negative Rate (TNR)}.

\para{Results: Failure Monitoring Performance} Both \vlmdynlatentbin and \classdynlatent perform well when evaluation task description is the same as training. However, their performances drop sharply in novel task description, indicating poor generalization of end-to-end model even though they have the same model capacity as our method.  In contrast, \ours consistently attains high accuracy, higher than $75\%$, with balanced TPR and TNR across all tasks given different task descriptions, demonstrating its reliability and flexibility as a failure monitor.

The generalization capability of our method comes from decoupling the problem of failure monitoring as behavior narration and outcome evaluation. 

This is also demonstrated by \vlmact, which has the same intermediate step and shows balanced performance in both task descriptions, but \vlmact often misclassifies behaviors in \textit{Bag Task} as failures, resulting in low TPR and lower overall accuracy than \ours.

 In Fig.~\ref{fig:policy_monitoring_more}, we demonstrate monitoring results for all three different behaviors qualitatively. Across all three modes of behaviors, \ours consistently generates accurate descriptions as well as correct monitoring results. \vlmact is biased to generate failure narrations across all three modes, leading to wrong monitoring results. \classdynlatent and \vlmdynlatentbin completely do not understand different action plans within different task descriptions. They generate the same monitoring results for totally different descriptions, contradictory to the actual execution.

\para{Component-level Analysis for the System}
We analyze each component in our method as well as all the baselines in (Table~\ref{tab:policy_breakdown_all}). 
\ours shows high accuracy across all the components while the poor performance of \vlmact is from the low-quality of narration directly generated from low-level action sequences. After finetuning, both methods preserve the strong reasoning capability of the VLM. However, \classdynlatent and \vlmdynlatentbin has a huge performance drop in novel task scenarios because they are overfitting to the specific task description in the training.
\\

\subsubsection{Metric Ablations}\hfill 
\label{sec:appendix_ablations}

\begin{figure}[h!]
   \vspace{-0.3cm}
    \centering
    \includegraphics[width=0.8\linewidth]{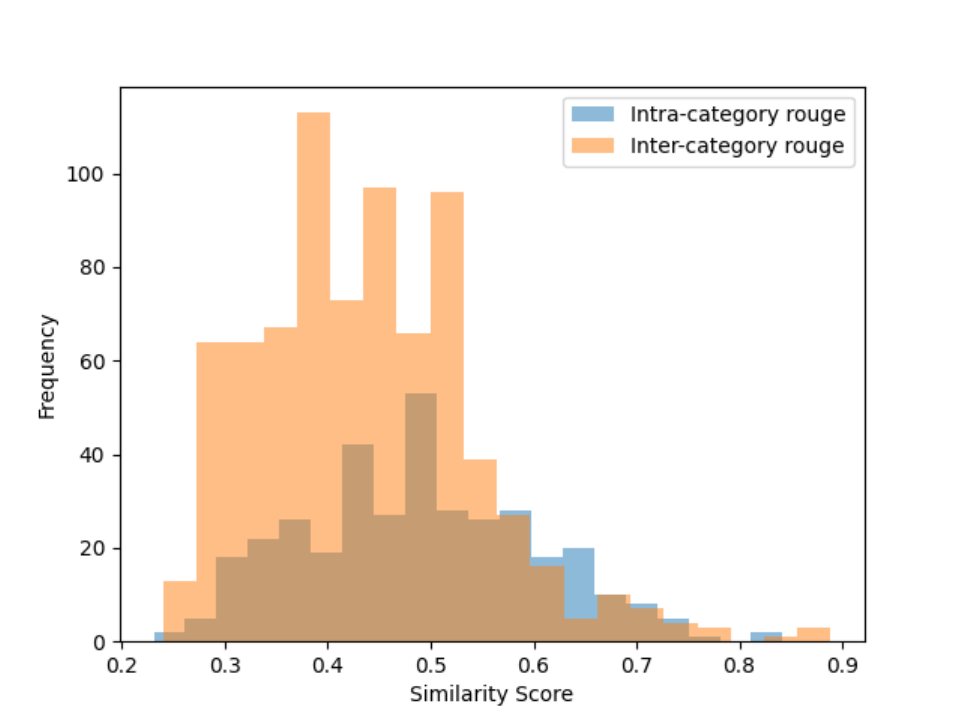}
    \caption{\textbf{Distribution of ROUGE-L Score} shows that intra-category scores are overlapped with inter-category ones.}
    \label{fig:rouge_score}
    \vspace{-0.3cm}
\end{figure}

\begin{figure}[h!]
\vspace{-0.5cm}
    \centering
    \includegraphics[width=0.8\linewidth]{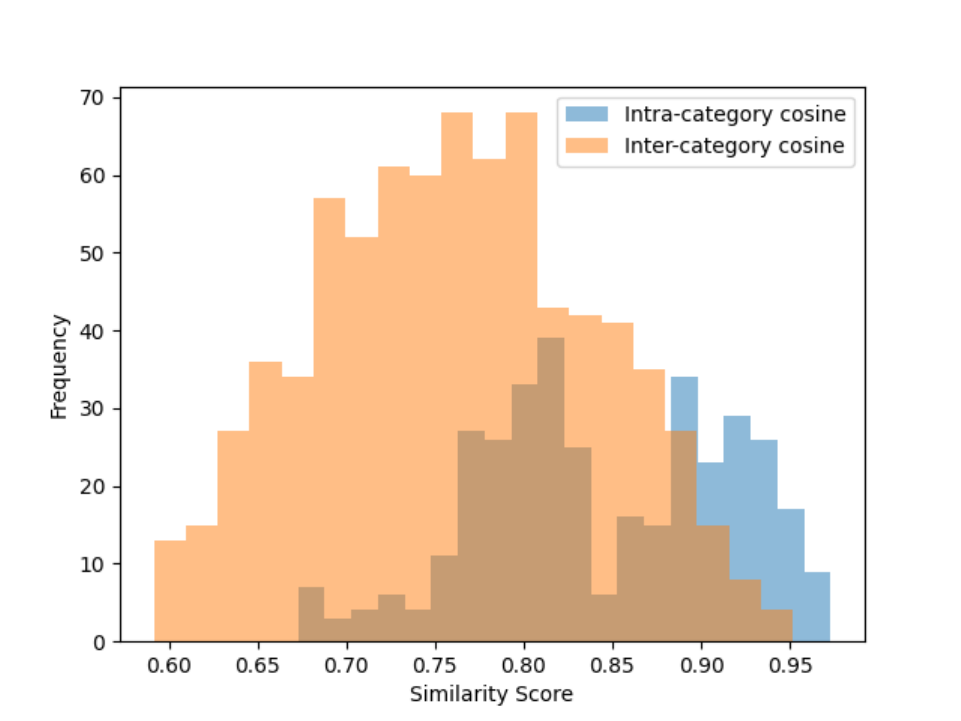}
    \caption{\textbf{Distribution of Cosine Similarity Score} shows that intra-category scores are overlapped with inter-category ones.}
    \label{fig:cosine_score}
    \vspace{-0.4cm}
\end{figure}

\begin{figure}[h!]
    \centering
    \vspace{-0.3cm}
    \includegraphics[width=0.8\linewidth]{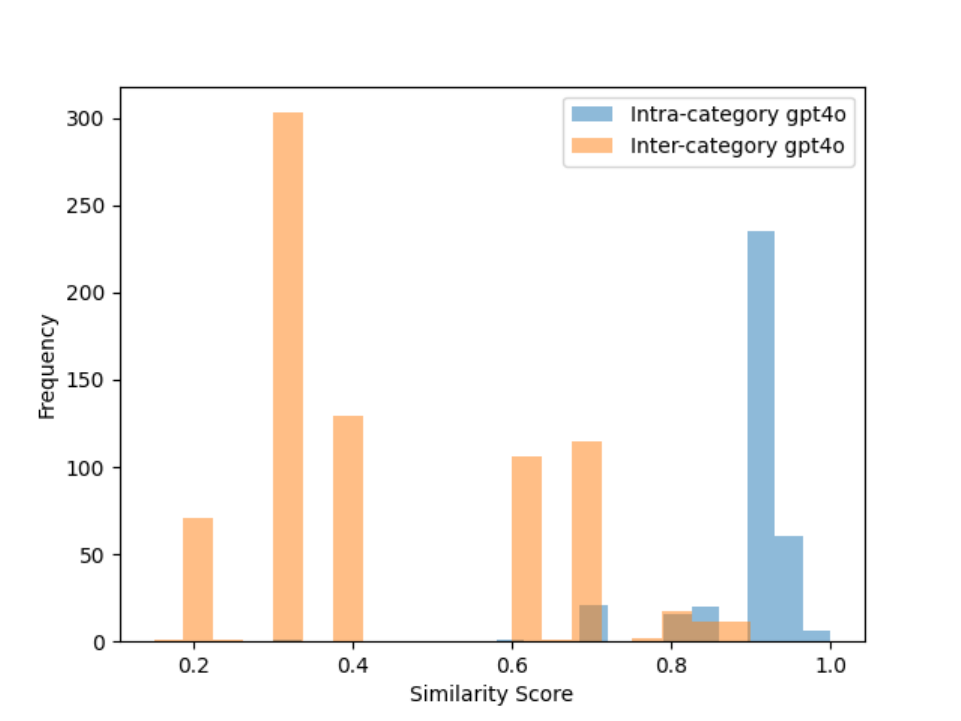}
    \caption{\textbf{Distribution of LLM Score} shows inter-category  and inter-category scores can be roughly separated at 0.7.}
    \label{fig:llm_score}
\end{figure}

\para{Metrics for Behavior Narration}
We investigate four common text-generation metrics proposed in prior work~\citep{duanaha}: \textit{Cosine Similarity}, \textit{ROUGE-L}, \textit{LLM Fuzzy Matching}, and \textit{Binary Success Rate}. To assess each metric’s correlation with ground-truth labels, we sample 16 narrations for each of three behaviors in the \textbf{Cup Task} (grasping by handle, grasping by rim, and grasp failures), yielding 360 intra-category and 768 inter-category comparisons. As shown in Figures.~\ref{fig:rouge_score},\ref{fig:cosine_score}, and\ref{fig:llm_score}, it is difficult for Cosine Similarity and ROUGE-L to cleanly distinguish narrations from the same category versus different categories, whereas \textit{LLM Fuzzy Matching} with GPT-4o can easily separate them. This discrepancy arises because the narrations share similar high-level semantics (e.g., “grasping the cup”) and differ only in fine-grained details (e.g., grasp location). Consequently, we adopt \textbf{LLM Score} and \textbf{Ground-Truth Accuracy} (manual matching) as our final metrics for evaluating generated behavior.

\end{document}